\def\year{2022}\relax
\documentclass[letterpaper]{article} % DO NOT CHANGE THIS
\usepackage{aaai22}  % DO NOT CHANGE THIS
\usepackage{times}  % DO NOT CHANGE THIS
\usepackage{helvet}  % DO NOT CHANGE THIS
\usepackage{courier}  % DO NOT CHANGE THIS
\usepackage[hyphens]{url}  % DO NOT CHANGE THIS
\usepackage{graphicx} % DO NOT CHANGE THIS
\usepackage{natbib}  % DO NOT CHANGE THIS AND DO NOT ADD ANY OPTIONS TO IT
\usepackage{caption} % DO NOT CHANGE THIS AND DO NOT ADD ANY OPTIONS TO IT
\DeclareCaptionStyle{ruled}{labelfont=normalfont,labelsep=colon,strut=off} % DO NOT CHANGE THIS
\usepackage{algorithm}
\usepackage{algorithmic}

\usepackage{newfloat}
\usepackage{listings}

\usepackage{xspace}
\usepackage{times}
\usepackage{latexsym}
\usepackage{tabularx}

\newcommand{\syntdec}{\texttt{SyntDEC}\xspace}
\newcommand{\syntdecmean}{\texttt{SyntDEC\textunderscore Mean}\xspace}
\newcommand{\syntdecmorph}{\texttt{SyntDEC\textunderscore Morph}\xspace}

\newcommand{\syntdecmorphparsed}{\texttt{SyntDEC\textunderscore Morph (Parsed)}\xspace}

\newcommand{\dec}{\texttt{DEC}\xspace}
\newcommand{\sae}{\texttt{SAE}\xspace}
\newcommand{\kmeansmethod}{\texttt{KMeans}\xspace}

\newcommand{\diora}{\texttt{DIORA}\xspace}
\newcommand{\spanbert}{\texttt{SpanBERT}\xspace}
\newcommand{\elmo}{\texttt{ELMo}\xspace}

\newcommand{\fasttext}{\texttt{fastText}\xspace}
\newcommand{\cbow}{\texttt{CBoW}\xspace}
\newcommand{\colab}{\texttt{CoLab}\xspace}
\newcommand{\posi}{\texttt{POSI}\xspace}
\newcommand{\docl}{\texttt{DoClu}\xspace}
\newcommand{\mbert}{\texttt{mBERT}\xspace}
\newcommand{\ebert}{\texttt{E-BERT}\xspace}
\newcommand{\debert}{\texttt{DeBERT}\xspace}

\newcommand{\explayer}{\texttt{Expected Layer}\xspace}
\usepackage{multirow}
\usepackage{comment}
\usepackage{sidecap}
\usepackage{soul}
\usepackage{amsfonts}
\usepackage{amsmath}
\usepackage{amssymb}
\usepackage{nicefrac}
\usepackage{booktabs} % Pretty tables
\usepackage{makecell} % cell line breaks..
\usepackage{graphicx}
\usepackage[inline]{enumitem}
\usepackage[font=small]{subcaption}
\usepackage[font=small]{caption}
\usepackage{xargs} % commandx
\usepackage[autostyle]{csquotes}  

\usepackage[T1]{fontenc}

\usepackage{color}
\usepackage{soul}
\usepackage[textsize=scriptsize]{todonotes}
\definecolor{pigment}{rgb}{0.2, 0.2, 0.6}

\definecolor{blue}{RGB}{0, 93, 170}			%Go Big Blue!
\definecolor{darkgreen}{HTML}{3bb35b}

\definecolor{darkyellow}{HTML}{f0b71d}

\newcommand{\ignore}[1]{}
\usepackage{multirow}
\usepackage{float}

\floatstyle{ruled}
\newfloat{listing}{tb}{lst}{}
\floatname{listing}{Listing}
\title{Deep Clustering of Text Representations for Supervision-free Probing of Syntax}
\author{
    Vikram Gupta,\textsuperscript{\rm 1}
    Haoyue Shi,\textsuperscript{\rm 2}
    Kevin Gimpel,\textsuperscript{\rm 2} 
    Mrinmaya Sachan\textsuperscript{\rm 3}
}
\affiliations{
    \textsuperscript{\rm 1} ShareChat, India \\
    \textsuperscript{\rm 2} Toyota Technological Institute at Chicago, IL, USA\\
    \textsuperscript{\rm 3} Department of Computer Science, ETH Zurich \\
    vikramgupta@sharechat.co, freda@ttic.edu, kgimpel@ttic.edu, mrinmaya.sachan@inf.ethz.ch
}

\usepackage{bibentry}
\begin{document}

\maketitle

\begin{abstract}
	We explore deep clustering of text representations for unsupervised model interpretation and induction of syntax. As these representations are high-dimensional, out-of-the-box methods like \kmeansmethod do not work well. Thus, our approach jointly transforms the representations into a lower-dimensional cluster-friendly space and clusters them. We consider two notions of syntax: \textit{part of speech induction} (\posi) and \textit{constituency labelling} (\colab) in this work. Interestingly, we find that Multilingual BERT (\mbert) contains surprising amount of syntactic knowledge of English; possibly even as much as English BERT (\ebert). 
	Our model can be used as a supervision-free probe which is arguably a less-biased way of probing. We find that unsupervised probes show benefits from higher layers as compared to supervised probes.
	We further note that our unsupervised probe utilizes \ebert and \mbert representations differently, especially for \posi.
	We validate the efficacy of our probe by demonstrating its capabilities as a unsupervised syntax induction technique.
	Our probe works well for both syntactic formalisms by simply adapting the input representations. We report competitive performance of our probe on 45-tag English \posi, state-of-the-art performance on 12-tag \posi across 10 languages, and competitive results on \colab. We also perform zero-shot syntax induction on resource impoverished languages and report strong results.
	%\footnote{Source code will be made available upon acceptance.}
	%find improvements by incorporating small amounts of task-specific information as well as by 
	
	%\vikram{We also show that the approach can be used in cross-lingual learning as the models learnt on a resource-rich language can be used for languages with less data. 
	%Moreover, the approach can also be seen as a supervision-free probe to discover linguistic structure in contextualized text representations.
	%}
	%With this work, we establish strong \vikram{baselines for unsupervised \posi, cross-lingual \posi and \colab} using contextualized text representations.

	%\kevin{It would be nice to state here some of our most exciting findings, like the consistent improvement with mBERT over E-BERT even for English tasks, the strong performance of crosslingual POSI and how it works best for more closely related languages, and some findings from the analysis: for unsup POS, higher expected layers than sup POS, higher expected layers for mBERT than English BERT, etc. we will have to pick and choose, but these things are prob at least as exciting for readers than SOTA POSI results...}
	
\end{abstract}
\section{Introduction}
\label{sec:intro}
%\kevin{A general thing: I think we should be careful about using the term ``cluster friendly space'' because we don't actually know if the space is more friendly to clustering (unless we've established that quantitatively somehow)}
%NLP has seen a paradigm shift in recent years with the adoption of self-supervised contextualized representations of text \cite[\emph{inter alia}]{peters2018deep,devlin2018bert}. 
Contextualized text representations \cite{peters2018deep,devlin2018bert} have been used in many supervised NLP problems such as part-of-speech (POS) tagging \cite{tsai2019small}, syntactic parsing \cite{kitaev2018constituency,zhou2019head,mrini2019rethinking}, and coreference resolution \cite{lee2018higher,joshi2019bert,wu-etal-2020-corefqa}, often leading to significant improvements.
Recent works have shown that these representations encode linguistic information including POS \cite{belinkov2017neural}, morphology \cite{peters2018deep}, and syntactic structure \cite{linzen2016assessing,peters2018dissecting,tenney-etal-2019-bert,hewitt2019structural}. 
%These methods work by specifying a probe, a supervised model for finding information in the language representation.

While there has been a lot of focus on using contextualized representations in supervised settings for either solving NLP problems and interpreting these representations, the efficacy of these representations for unsupervised learning is not well explored\footnote{
Some recent work such as DIORA \cite{drozdov2019unsupervised1,drozdov2019unsupervised2} has explored specialized methods for unsupervised discovery and representation of constituents using ELMo \cite{peters2018deep}. 
%by incorporating the inside-outside algorithm into a latent tree chart parser.
\cite{jin2019unsupervised} used ELMo with a normalizing flow model while \cite{cao2020unsupervised} used RoBERTa~\cite{liu2019roberta} for unsupervised constituency parsing.}. Most of the recent work in ``probing'' contextual representations have focused on building supervised classifiers and using accuracy to interpret these representations.  This has led to a debate as it is not clear if the supervised probe is probing the model or trying to solve the task \cite{hewitt2019structural,pimentel-etal-2020-information}. 

Thus, in this work, we explore a new clustering-based approach to probe contextualized text representations.
Our probe allows for studying text representations with relatively less task-specific transformations due to the absence of supervision. Thus, our approach is arguably a less biased way to discover linguistic structure than supervised probes \cite{hewitt2019structural,pimentel-etal-2020-information,zhou-21}.%\vikram{ 
We focus on two syntactic formalisms: part-of-speech induction (\posi) and constituency labelling (\colab), and explore the efficacy of contextualized representations towards encoding syntax in an unsupervised manner.
We investigate the research question: \textit{Do contextualized representations encode enough information for unsupervised syntax induction? How do these perform on \posi, which has been traditionally solved using smaller context windows and morphology and span-based \colab?}

\begin{figure}
	\centering
	\begin{subfigure}[t]{.485\columnwidth}
		\centering
		\includegraphics[width=\textwidth]{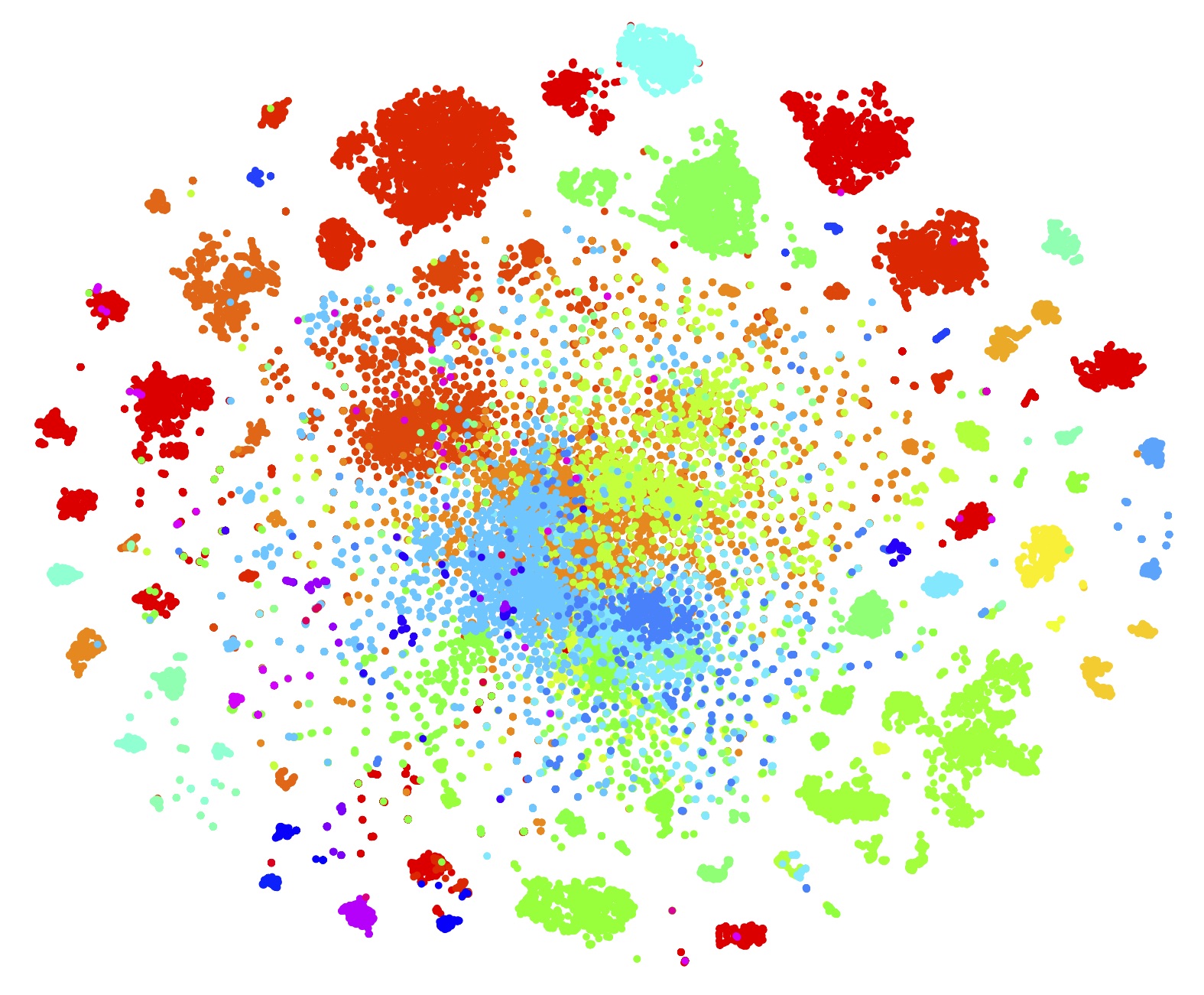}
		\caption{\mbert}
	\end{subfigure}
	\begin{subfigure}[t]{.485\columnwidth}
		\centering
		\includegraphics[width=\textwidth]{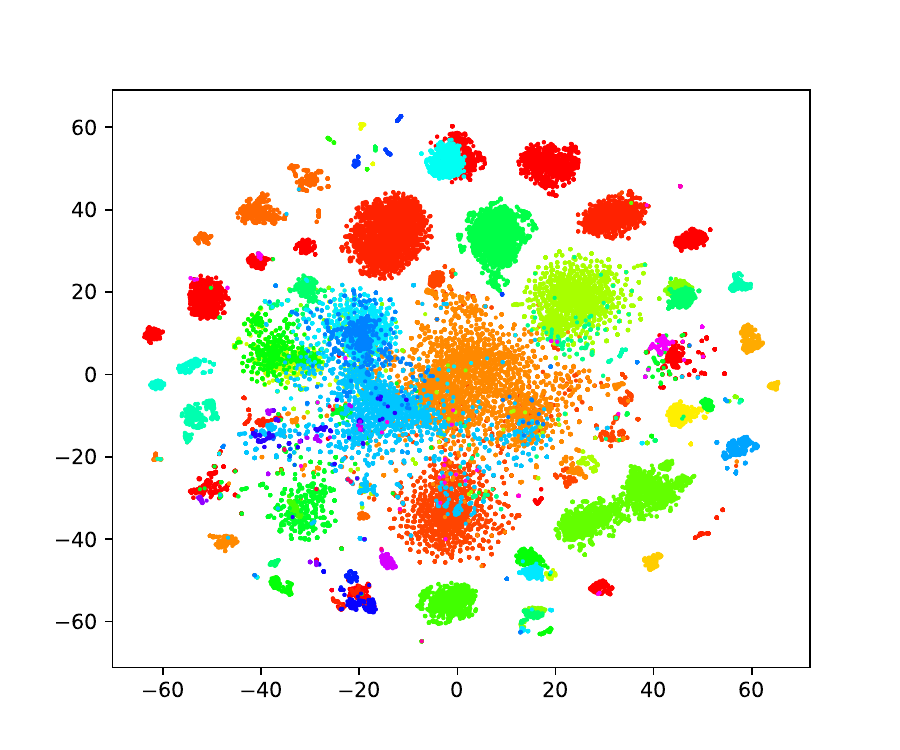}
		\caption{\syntdec}
	\end{subfigure}
	\caption{t-SNE visualization of \mbert embeddings (clustered using kMeans) and \syntdec (our probe) embeddings of tokens from Penn Treebank. Colors correspond to ground truth POS tags.}\label{fig:tsne}
\end{figure}

For both formalisms, we find that naively clustering text representations does not perform well.  We speculate that this is because contextualized text representations are high-dimensional and not very friendly to existing clustering approaches. Thus, we develop a deep clustering approach \cite{xie2016unsupervised,ghasedi2017deep,jiang2016variational,chang2017deep,yang2016joint,yang2017towards} which transforms these representations into a lower dimensional,  clustering friendly latent space.
This transformation is learnt jointly with the clustering using a combination of reconstruction and clustering objectives. The procedure iteratively refines the transformation and the clustering using an auxiliary target distribution derived from the current soft clustering. As this process is repeated, it gradually improves the transformed representations as well as the clustering.
We show a t-SNE visualization of \mbert embeddings and embeddings learned by our deep clustering probe (\syntdec) in Figure \ref{fig:tsne}. 

We further explore architectural variations such as pretrained subword embeddings from \fasttext \cite{joulin2016bag}, a continuous bag of words (\cbow) loss \cite{mikolov2013efficient}, and span representations
\cite{toshniwal2020crosstask} to incorporate task-dependent information into the latent space and observe sugnificant improvements.
It is important to note that we do not claim that clustering contextualized representations is the optimal approach for \posi as representations with short context~\cite{lin-etal-2015-unsupervised},~\cite{he2018unsupervised} and word-based \posi~\cite{yatbaz2012learning} have shown best results. Our approach explores the potential of contextualized representations for unsupervised induction of syntax and acts as an unsupervised probe for interpreting these representations. Nevertheless, we report competitive many-to-one (M1) accuracies for \posi on the 45-tag Penn Treebank WSJ dataset as compared to specialized state-of-the-art approaches in the literature \cite{he2018unsupervised} and improve upon the state of the art on the 12 tag universal treebank dataset across multiple languages \cite{stratos2016unsupervised,stratos2018mutual}. 
We further show that our approach can be used in a zero-shot crosslingual setting where a model trained on  one  language  can used for evaluation in another language. We observe impressive crosslingual \posi performance, showcasing the representational power of \mbert, especially when the languages are related. Our method also achieves competitive results on \colab on the WSJ test set, outperforming the initial \diora approach \cite{drozdov2019unsupervised1} and performing comparably to recent DIORA variants \cite{drozdov2019unsupervised2} which incorporate more complex methods such as latent chart parsing and discrete representation learning. In contrast to specialized state-of-the-art methods for syntax induction, our framework is more general as it demonstrates good performance for both \colab and \posi by simply adapting the input representations.

We further investigate the effectiveness of multilingual BERT  (\mbert) \cite{devlin2018bert}
for \posi across multiple languages and \colab in English and see improvement in performance by using \mbert for both tasks even in English. This is in contrast with the supervised experiments where both \mbert and \ebert perform competitively.
%A possible explanation could be that since mBERT looks at multiple languages, it is encouraged to encode generic information across languages which helps under unsupervised settings.} 
 In contrast to various supervised probes in the literature \cite{liu-etal-2019-linguistic,tenney-etal-2019-bert},
our unsupervised probe finds that syntactic information is captured in higher layers on average than what was previously reported 
%by supervised probes in the past 
\cite{tenney-etal-2019-bert}. 
Upon further layer-wise analysis of the two probes, we find that while supervised probes show that all layers of \ebert contain syntactic information fairly uniformly, middle layers lead to a better performance on the investigated syntactic tasks with our unsupervised probe. 
% the unsupervised probe shows that the highest layers contain less syntactic information than the middle layers.
%}

%\kevin{Rather than say our results are not that different, it would be more interesting to frame this as ``Here's what's different'' and then give our two favorite findings from the probing section}
%Findings from our unsupervised probe mostly align with 
%However, we also observe a few small differences with supervised probes which warrant a deeper investigation. For instance, 
%\textcolor{red}{
%that for both \posi and \colab, \mbert demonstrates higher center of gravity (COG) as compared to \ebert. We hypothesize that initial layers of \mbert are involved in processing different languages, thus delaying the syntactic understanding to later layers.  We also find that the COG is higher for supervised probes as compared to unsupervised ones.
%Higher COG under supervised settings can be attributed to evenly distributed weights across different layers unlike unsupervised, where weights are peaky. 

\begin{figure}
\centering
\includegraphics[width=\columnwidth]{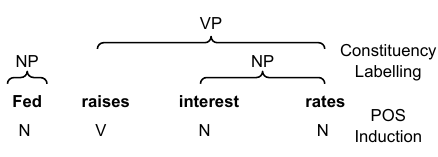}
\caption{
An illustration of \posi and \colab formalisms.
}
\label{pos_colab}
\end{figure}

\section{Problem Definition}\label{sec:task}
%We examine the problem of inducing syntax in a completely unsupervised way.
%from contextualized word representations.
%Specifically, i
%In this work, w
We consider two syntax induction problems in this work:
\begin{enumerate}[noitemsep,nolistsep]
    \item {\bf Part-of-speech induction (\posi)}: determining part of speech of words in a sentence. 
    %\freda{Sorry for bringing these abbr. things again -- I felt it's better to have a separate name for our task and previous mentions of POS, so I changed it to POSI.}
    \item {\bf Constituency label induction (\colab)}: determining the constituency label for a given constituent (span of contiguous tokens).\footnote{Note that it is not necessary for constituents to be contiguous, but we only consider contiguous constituents for simplicity.}
    %\kevin{I changed ``connected'' to ``consecutive''. Hopefully I didn't change the meaning you had intended.}
    % \freda{A constituency is not necessary to be always connected, so I added a footnote here.}
\end{enumerate} 
Figure~\ref{pos_colab} shows an illustration for the two tasks. In order to do well, both tasks require reasoning about the context. This motivates us to use contextualized representations, which have shown an 
%amazing 
ability to model such information effectively. Letting $[m]$ denote $\{1, 2, \ldots, m\}$, we model unsupervised syntax induction as the task of learning a mapping function $C: X \longrightarrow [m]$. For \posi, $X$ is the set of word tokens in the corpus and $m$ is the number of part-of-speech tags.\footnote{$X$ is distinct from the corpus vocabulary; in \posi, we tag each word token in each sentence with a POS tag.}
For \colab, $X$ is the set of constituents across all sentences in the corpus and $m$ is the number of constituent labels. For each element $x \in X$, 
%\footnote{Word tokens for POS tagging, constituents for CoLab.} 
let $c(x)$ denote the context of $x$ in the sentence containing $x$.
The number $m$ of true clusters is assumed to be known. 
%, i.e., the number of POS tags for the POSI task and the number of constituent labels for the CoLab task. 
For \colab, we also assume gold constituent spans from manually annotated constituency parse trees, focusing only on determining constituent labels, following \citet{drozdov2019unsupervised2}.

%%\begin{figure}
%%\includegraphics[width=%%\columnwidth]{sections/%%images/DEC-Training.pdf%%}
%%\caption{An illustration of our SyntDEC model.}\label{syntdec}
%%\end{figure}

\section{Proposed Method}
We address unsupervised syntax induction via clustering, where $C$ defines a clustering of $X$ into $m$ clusters.
We define a deep embedded clustering framework and modify it to support common NLP objectives such as continuous bag of words \citep[]{mikolov2013efficient}. Our framework jointly transforms the text representations into a lower-dimensions and learns the clustering parameters in an end-to-end setup. 
%We begin with a brief summary of DEC \cite{xie2016unsupervised} (a popular deep clustering approach) below.

% The proposed method addresses unsupervised syntax induction as a clustering problem where the number of clusters are set to the number of ground truth tags and the representations are extracted from pretrained networks. In traditional clustering, these representations are assumed to be fixed and the clustering parameters are learned using clustering algorithms like KMeans. However, in this work, we transform the pretrained representations into cluster friendly lower dimension spaces along with learning clustering parameters in an end-to-end setup.

\subsection{Deep Clustering}
\label{sec:DEC}
%Our framework is inspired by previous works in deep clustering.
Unlike traditional clustering approaches that work with fixed, and often hand-designed features, deep clustering \citep{xie2016unsupervised,ghasedi2017deep,jiang2016variational,chang2017deep,yang2016joint,yang2017towards} transforms the data $X$ into a latent feature space $Z$ with a mapping function $f_\theta : X \longrightarrow Z$, where $\theta$ are learnable parameters. The dimensionality of $Z$ is typically much smaller than $X$. The datapoints are clustered by simultaneously learning a clustering $\Tilde{C}: Z \rightarrow [m]$.While $C$  might have been hard to learn directly (due to the high dimensionality of $X$), learning $\Tilde{C}$ may be easier.

\begin{figure}
\centering
\includegraphics[width=0.7\columnwidth]{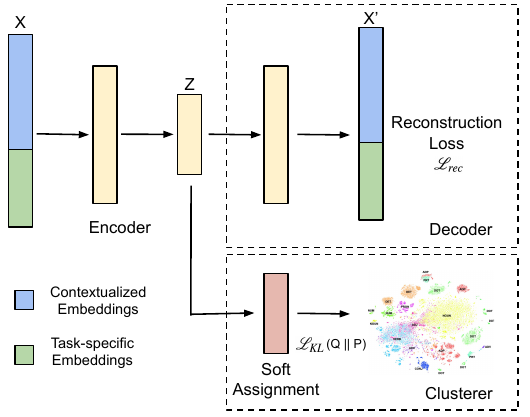}
\caption{An illustration of our \syntdec model.}\label{syntdec}
\end{figure}
\noindent{\bf Deep Embedded Clustering:}
We draw on a particular deep clustering approach: Deep Embedded Clustering \citep[\dec;][]{xie2016unsupervised}. Our approach consists of two stages: (a) a \emph{pretraining} stage, and (b) a \emph{joint representation learning and clustering} stage. 
% The pretraining step 
%\begin{figure}
%\includegraphics[trim=0 0 120 0,clip,width=\columnwidth]{sections/images/DEC-Training.pdf}
%\caption{SyntDEC architecture}\label{syntdec}
%\end{figure}
In the \textit{pretraining} stage, a mapping function $f_\theta$ is pretrained using a stacked autoencoder (\sae). 
The \sae learns to reconstruct $X$ through the bottleneck $Z$, i.e., $X \xrightarrow[]{\mathit{encoder}} Z \xrightarrow[]{\mathit{decoder}} X'$.
We use mean squared error (MSE) as the reconstruction loss:
\begin{align}
    \mathcal{L}_{\mathit{rec}} = ||X-X'||^2
    = \sum\limits_{x\in X} ||x-x'||^2 \nonumber
\end{align}
% a denoising stacked autoencoder is learnt to transform the representations to a lower dimension latent space using the reconstruction loss. We use mean square loss (MSE) as the reconstruction loss. 
The encoder parameters are used to initialize the mapping function $f_\theta$.

In the \textit{joint representation learning and clustering} stage, we finetune the encoder $f_\theta$ trained in the pretraining stage to minimize a clustering loss $\mathcal{L}_\textit{KL}$. The goal of this step is to learn a latent space that is amenable to clustering.
% (decoder is removed in this stage). 
%There are a number of deep embedded clustering approaches in the literature \cite{ghasedi2017deep,jiang2016variational,chang2017deep,yang2016joint,yang2017towards} which differ in their choice of the clustering method. 
We learn a set of $m$ cluster centers $\{\mu_i \in Z\}_{i=1}^m$ of the latent space $Z$ and alternate between computing an \textit{auxiliary target distribution}
and minimizing the Kullback-Leibler (KL) divergence. First, a soft cluster assignment is computed for each embedded point. Then, the mapping function $f_\theta$ is refined along with the cluster centers by learning from the assignments using an auxiliary target distribution. This process is repeated.
%\freda{I moved the citation of ver der Maaten and Hinton to the position we mention t-SNE, since here we're talking about Student's t-distribution in general.}
The soft assignment is computed via the Student's $t$-distribution. 
% \cite{maaten2008visualizing}
The probability 
of assigning data point $i$ to cluster $j$ is denoted $q_{ij}$ and defined:
\begin{eqnarray}
q_{ij} = \frac{(1+||z_{i} - \mu_j||^2/\nu)^{-\frac{\nu+1}{2}}}{\sum\limits_{j'}(1+||z_i - \mu_{j'}||^2/\nu)^{-\frac{\nu+1}{2}}}\nonumber 
\end{eqnarray}
where $\nu$ is set to 1 in all experiments.
Then, a cluster assignment hardening loss \cite{xie2016unsupervised} is used to make these soft assignment probabilities more peaked. 
%\kevin{I changed the $P$ and $Q$ to $p$ and $q$ because I think that's more common when talking about distributions; I think the DEC paper's use of capital letters for distributions is a little confusing} 
This is done by letting cluster assignment probability distribution $q$ approach a more peaked auxiliary (target) distribution $p$:
\begin{equation}
{p}_{ij} = \frac{q_{ij}^2/n_{j}}{\sum_{j'} q_{ij'}^2/n_{j'}} \quad\quad \quad {n}_{j} = \sum_{i}{q}_{ij} \nonumber
\end{equation}

\noindent By squaring the original distribution and then normalizing it, the auxiliary distribution $p$ forces assignments to have more peaked  probabilities. This aims to improve cluster purity, put emphasis on data points assigned with high confidence, and to prevent large clusters from distorting the latent space. The divergence between the two probability distributions is formulated as the Kullback-Leibler divergence:

\begin{equation}
    \mathcal{L}_\textit{KL} = \sum_i \sum_j p_{ij} \log \frac{p_{ij}}{q_{ij}} \nonumber 
\end{equation}
\noindent 
The representation learning and clustering model is learned end-to-end.
\subsection{\syntdec: DEC for Syntax Induction}
\label{sec:SyntDEC}

%In this work, 
We further modify \dec for syntax induction:

%\noindent{\bf a) CBoW and Skip-gram autoencoders:}
\noindent{\bf a) \cbow autoencoders:} While \dec uses a conventional autoencoder, i.e., the input and output are the same, we modify it to support the 
%common NLP objectives like 
%skip-gram and 
continuous bag of words (\cbow) objective %\citep[\cbow;][]{mikolov2013efficient} 
\citep{mikolov2013efficient}. This helps focus the low-dimensional representations to focus on context words, which are expected to be helpful for \posi. 
In particular, given a set of tokens $c(x)$ that defines the context for an element $x \in X$, 
%the Skip-gram model uses the representation of the element $x$ to predict words in the context $c(x)$ and 
 \cbow combines the distributed representations of tokens in $c(x)$ to predict the element $x$ in the middle. See Appendix \ref{cbow_illustration} for an illustration.

\noindent{\bf b) Finetuning with reconstruction loss:} We found that in the clustering stage, finetuning with respect to the KL divergence loss alone easily leads to trivial solutions where all points map to the same cluster. To address this, we add the reconstruction loss as a regularization term. This is in agreement with subsequent works in deep clustering \citep{yang2017towards}. Instead of solely minimizing $\mathcal{L}_\textit{KL}$, we minimize 
% We also note that for syntax induction, it is important to regularize the learning of the latent space with the reconstruction loss even in the clustering stage. In the absence of this regularization, the pretrained representations collapse. Thus, the effective loss for our framework is weighted average between the KL divergence loss and reconstruction loss. 
\begin{equation}
\mathcal{L}_\textit{total} = \mathcal{L}_\textit{KL} + \lambda\mathcal{L}_\textit{rec}
\end{equation}
in the clustering stage, where $\lambda$ is a hyperparameter denoting the weight of the reconstruction loss.

\noindent{\bf c) Contextualized representations:} We represent linguistic elements $x$ by embeddings extracted from pretrained networks like BERT \cite{devlin2018bert}, SpanBERT \cite{joshi:spanbert}, and multilingual BERT \cite{devlin2018bert}. All of these networks are multi-layer architectures. Thus, we average the embeddings across the various layers. We experimented with different layer combinations but found the average was the best solution for these tasks. We averaged the embeddings of the subword units to compute word embeddings.\footnote{In our preliminary experiments, we also tried other pooling mechanisms such as min/max pooling over subwords, but average performed the best among all of them.}
For \colab, we represent spans by concatenating the representations of the end points \cite{toshniwal2020crosstask}. 

\noindent{\bf d) Task-specific representations:} Previous work in unsupervised syntax induction has shown the value of task-specific features. In particular, a number of morphological features based on prefixes and suffixes and spelling cues like capitalization have been used in unsupervised \posi works \cite{tseng-etal-2005-morphological,stratos2018mutual, yatbaz2012learning}.
In our \posi experiments, we incorporate these morphological features by using word representations from \fasttext \citep{joulin2016bag}. 
%Suffixes of words like \textit{"playing", "grapes", "girl"} etc. are most informative for \posi. Thus, 
We use \fasttext embeddings of the trigram from each word with contextualized representations as input.
%We also tried unigram and bigram representations but observed trigram to be most effective.

\section{Experimental Details}

{\bf Datasets:}
We evaluate our approach for \posi on two datasets: 45-tag Penn Treebank Wall Street Journal (WSJ) dataset \citep{marcus1993building} and multilingual 12-tag datasets drawn from the universal dependencies project~\cite{nivre-etal-2016-universal}.
%\kevin{we should cite the universal dependencies paper.. i think it's Nivre et al with a ton of authors}.
The WSJ dataset has approximately one million words tagged with 45 part of speech tags. For multilingual experiments, we use the 12-tag universal treebank v2.0 dataset which consists of corpora from 10 languages.\footnote{We use v2.0 in order to compare to  \citet{stratos2018mutual}.} The words in this dataset have been tagged with 12 universal POS tags \cite{mcdonald2013universal}. For \colab, we follow the existing benchmark \cite{drozdov2019unsupervised2} and evaluate on the WSJ test set.
%\mrinmaya{Any links to the datasets?} 
%\freda{added citations, and I think it's hopefully easy to find them following them. }
For \posi, as per the standard practice \cite{stratos2018mutual}, we use the complete dataset (train + val + test) for training as well as evaluation. However, for \colab, we use the train set to train our model and the test set for reporting results, following \citet{drozdov2019unsupervised2}. 

\noindent{\bf Evaluation Metrics:}
For \posi, we use the standard measures of many-to-one \citep[M1;][]{johnson2007doesn} accuracy and V-Measure~\cite{vmeasure}. 
%\vikram{we have removed NMI from all tables now so we can remove NMI from here also.}
For \colab, we use F1 score following \citet{drozdov2019unsupervised2}, ignoring spans which have only a single word and spans with the ``TOP'' label. In addition to F1, we also report M1 accuracy for \colab to show the clustering performance more naturally and intuitively. 
% also for \vikram{We can explain motivation of reporting M1 for CoLab here?}

\noindent{\bf Training Details:}
Similar to \citet{xie2016unsupervised}, we use greedy layerwise pretraining \cite{bengio2007greedy} for initialization. New hidden layers are successively added to the autoencoder, and the layers are trained to denoise output of the previous layer. 
%The output of the previous layer is corrupted by applying dropout and is used as input and the layer is trained to reconstruct the original input using mean square error.
%\freda{I was a little confused by ``layer-wise pretraining'' -- layer of what, what's the encoder's architecture? maybe we can discuss a bit about this.}\vikram{Added some more details but needs more polish}
After layerwise pretraining, we train the autoencoder end-to-end and leverage the trained \syntdec encoder (Section \ref{sec:SyntDEC}).
K-Means is used to initialize cluster means and assignments.
\syntdec is trained end-to-end with the reconstruction and clustering losses. 
%\vikram{Do we need the next paragraph? We have already mentioned this in Section 3.}
%After training the auto-encoder, .
%to transform the latent embeddings into a clustering friendly space. The clustering layer consists of NXM dimension matrix, where N is the number of clusters and M is the dimension of the latent embeddings. We use the soft assignment loss for this stage of clustering. We observed that using only the t-student loss can cause the embeddings to collapse to trivial solutions, so we use add the reconstruction loss via decoder. Adding the reconstruction loss stabilizes the training and avoids degenerate solutions.
%\freda{I wonder the last two paragraphs could be polished a bit -- I was slightly lost at a first glance.}
More details are in the appendix.
%~\ref{hyperparameters}.
%\freda{Seems that the first sentence is not appearing at a perfect place, maybe moving it elsewhere?}
%\mrinmaya{I think we need to say what we mean by oracle experiments. How is stopping chosen otherwise done}

\section{Part of Speech Induction (\posi)}
\label{sec:experiments}
%\subsection{P}

\begin{table}[t]
\small
\centering
\resizebox{\columnwidth}{!}{
\begin{tabular}{|l|c|c|} 
\hline
Method & M1 & VM\\
\hline
\syntdecmorph & 79.5 ($\pm$0.9) & 73.9 ($\pm$0.7)\\
%\syntdecmorph & 78.3 ($\pm$0.7) & 72.8 ($\pm$0.5)\\
\syntdec & 77.6 ($\pm$1.5) & 72.5 ($\pm$0.9)\\
\sae & 75.3 ($\pm$1.4) & 69.9 ($\pm$0.9)\\
\kmeansmethod & 72.4 ($\pm$2.9)  & - \\
\hline
\citet{brown1992class} & 65.6 ($\pm$NA) & -\\
\citet{stratos2016unsupervised} & 67.7 ($\pm$NA) & -\\
\citet{berg2010painless} & 74.9 ($\pm$1.5) & -\\ 
\citet{blunsom2011hierarchical} & 77.5 ($\pm$NA) & 69.8\\ 
\citet{stratos2018mutual} & 78.1 ($\pm$0.8) & -\\ 
\citet{tran2016unsupervised} & 79.1 ($\pm$NA) & 71.7 ($\pm$NA)\\
\citet{yuret2014unsupervised} & 79.5 ($\pm$0.3)  & 69.1($\pm$2.7) \\
\hline
\citet{yatbaz2012learning} (word-based) &80.2 ($\pm$0.7)  & 72.1 ($\pm$0.4) \\
\citet{he2018unsupervised} & 80.8 ($\pm$1.3) & 74.1 ($\pm$0.7)\\

\hline
\end{tabular}
}
\caption{Many-to-one (M1) accuracy and V-Measure (VM) of \posi on the 45-tag Penn Treebank WSJ dataset for 10 random runs. \mbert is used in all of our experiments (upper part of the table).}
\label{tab:temps}
\end{table}

\begin{table*}[h]

%\scriptsize
%\small
\footnotesize
\begin{center}
\resizebox{\textwidth}{!}{
\begin{tabular}{|l|c|c|c|c|c|c|c|c|c|c|c|} 
 %\hline
  \cline{2-12}
 \multicolumn{1}{c|}{} & de & en & es & fr & id & it & ja & ko & pt-br & sv & Mean \\ [0.5ex] 
 \hline%\hline
 \multirow{2}{*}{\sae} & 74.8 & 70.7 & 71.1 & 66.7 & 75.4 & 66.2 & 82.1 & 65.4 & 75.1 & 61.6 & 70.9 \\
 & ($\pm$1.5) & ($\pm$2.2) & ($\pm$2.4) & ($\pm$1.9) & ($\pm$1.6) & ($\pm$3.3) & ($\pm$0.9) & ($\pm$1.7)  & ($\pm$4.1) & ($\pm$2.6) &  \\\hline
\multirow{2}{*}{\syntdec} & \textbf{81.5} & \textbf{76.5} & \textbf{78.9} & \textbf{70.7} & \textbf{76.8} & \textbf{71.7} & \textbf{84.7} & \textbf{69.7} & \textbf{77.7} & \textbf{68.8} & \textbf{75.7} \\
& ($\pm$1.8) & ($\pm$1.1) & ($\pm$1.9) & ($\pm$3.9) & ($\pm$1.1) & ($\pm$3.3) & ($\pm$1.2) & ($\pm$1.5) & ($\pm$2.1) & ($\pm$3.9) & \\\hline
\hline
% \multirow{2}{*}{\syntdeor} & 81.7 & 76.7 & 79.5 & 70.8 & 76.9 & 71.8 & 84.7 & 69.7 & 78.9 & 69.7 & 76.0\\
% & ($\pm$1.8) & ($\pm$1.0) & ($\pm$1.7) & ($\pm$4.0) & ($\pm$1.2) & ($\pm$3.9) & ($\pm$1.2) & ($\pm$1.5) & ($\pm$1.8) & ($\pm$2.8) & \\
 %\hline\hline
  \multirow{2}{*}{\citet{stratos2018mutual}} & ~75.4 & 73.1 & 73.1 & 70.4 & 73.6 & 67.4 & 77.9 & 65.6 & 70.7 & 67.1 & 71.4\\
  & ($\pm$1.5) & ($\pm$1.7) & ($\pm$1.0) & ($\pm$2.9) & ($\pm$1.5) & ($\pm$3.3) & ($\pm$0.4) & ($\pm$1.2) & ($\pm$2.3) & ($\pm$1.5) & \\\hline
  \citet{stratos2016unsupervised} & ~63.4 & 71.4 & 74.3 & 71.9 & 67.3 & 60.2 & 69.4 & 61.8 & 65.8 & 61.0 & 66.7\\\hline
  \multirow{2}{*}{\citet{berg2010painless}} & ~67.5 & 62.4 & 67.1 & 62.1 & 61.3 & 52.9 & 78.2 & 60.5 & 63.2 & 56.7 & 63.2\\
  & ($\pm$1.8) & ($\pm$3.5) & ($\pm$3.1) & ($\pm$4.5) & ($\pm$3.9) & ($\pm$2.9) & ($\pm$2.9) & ($\pm$3.6) & ($\pm$2.2) & ($\pm$2.5) & \\\hline
  \citet{brown1992class} & 60.0 & 62.9 & 67.4 & 66.4 & 59.3 & 66.1 & 60.3 & 47.5 & 67.4 & 61.9 & 61.9\\
  \hline
\end{tabular}
}
\caption{M1 accuracy and standard deviations on the 12-tag universal treebank dataset averaged over 5 random runs. \mbert is used for all of our experiments (upper part of the table). The number of epochs are proportional to the number of samples and the M1 accuracy corresponding to the last epoch is reported.}
%For the oracle experiments, the models are allowed to run until the M1 accuracy saturates.
\label{table:2}
\end{center}
\end{table*}
\noindent{\bf 45-Tag Penn Treebank WSJ:}
In Table~\ref{tab:temps}, we evaluate the performance of contextualized representations and our probe on the 45-tag Penn Treebank WSJ dataset. \kmeansmethod clustering over the \mbert embeddings improves upon Brown clustering~\cite{brown1992class} (as reported by \citeauthor{stratos2018mutual}, \citeyear{stratos2018mutual}) and Hidden Markov Models  \cite{stratos2016unsupervised} based approach, showing that \mbert embeddings encode syntactic information.  The stacked autoencoder, \sae(trained during pretraining stage), improves upon the result of \kmeansmethod by nearly 3 points, which demonstrates the effectiveness of transforming the \mbert embeddings to lower dimensionality using an autoencoder before clustering. Our method (\syntdec) further enhances the result and shows that transforming the pretrained \mbert embeddings using clustering objective helps to extract syntactic information more effectively. When augmenting the \mbert embeddings with morphological features (\syntdecmorph), we improve over~\citet{stratos2018mutual} and ~\cite{tran2016unsupervised}. We also obtain similar M1 accuracy with higher VM as compared to~\cite{yuret2014unsupervised}. 

\noindent{\bf Morphology:}
We also note that the M1 accuracy of \citet{tran2016unsupervised} and \citet{stratos2018mutual} drop significantly by nearly 14 points in absence of morphological features, while \syntdec degrades by 2 points. This trend suggests that \mbert representations encode the morphology to some extent.

~\citet{yatbaz2012learning} are not directly comparable to our work as they performed word-based \posi which attaches same tag to all the instances of the word, while all the other works in Table~\ref{tab:temps} perform token-based \posi. They use task-specific hand-engineered rules like presence of hyphen, apostrophe etc. which might not translate to multiple languages and tasks. \cite{he2018unsupervised} train a \posi specialized model with Markov syntax model and short-context word embeddings and report current SOTA on \posi. In contrast to their method, \syntdec is fairly task agnostic.

\noindent{\bf 12-Tag Universal Treebanks:}
In Table~\ref{table:2}, we report M1 accuracies on the 12-tag datasets averaged over 5 random runs. Across all  languages, we report SOTA results and find an improvement on average over the previous best method \citep{stratos2018mutual} from 71.4\% to 75.7\%. We also note improvements of \syntdec over \sae (70.9\% to 75.7\%) across languages, which reiterates the importance of finetuning representations for clustering.
%Note that using contextualized representations from BERT-like models leads 
Our methods yield larger gains on this coarse-grained 12 tag \posi task as compared to the fine-grained 45 tag \posi task, and we hope to explore the reasons for this in future work.

\noindent{\bf Ablation Studies:}
Next, we study the impact of our choices on the 45-tag WSJ dataset. 
Table~\ref{table:3} demonstrates that multilingual BERT (\mbert) is better than English BERT (\ebert) across settings. 
For both \mbert and \ebert, compressing the representations with \sae and finetuning using \syntdec performs better than \kmeansmethod. Also, focusing the representations on the local context  (\cbow) improves performance with \ebert, though not with \mbert. In the appendix, we show the impact of using different types of \fasttext  character embeddings and note the best results when we use embeddings of the last trigram of each word.
\begin{table}[t]
\footnotesize
\begin{center}
\begin{tabular}{ |c|l|c| } 
%\hline
\cline{2-3}
\multicolumn{1}{c|}{}&Method & M1\\
\cline{2-3}
\hline
\parbox[t]{1.7mm}{\multirow{5}{*}{\rotatebox[origin=c]{90}{{\bf \ebert}}}}&\kmeansmethod & 69.1 ($\pm$0.9) \\\cline{2-3}
&\sae & 71.6 ($\pm$2.3) \\
&\cbow & 73.8 ($\pm$0.7) \\\cline{2-3}
&\syntdec (\sae) & 72.7 ($\pm$1.2) \\
&\syntdec (\cbow) & {\bf 74.4} ($\pm$0.6) \\
 \hline\hline
 \parbox[t]{1.7mm}{\multirow{5}{*}{\rotatebox[origin=c]{90}{{\bf \mbert}}}}&\kmeansmethod & 72.4 ($\pm$2.9) \\\cline{2-3}
&\sae & 75.3 ($\pm$1.4) \\ 
&\cbow & 75.1 ($\pm$0.3) \\\cline{2-3}
&\syntdec (\sae) & {\bf 77.8} ($\pm$1.4) \\ 
&\syntdec (\cbow) & {75.9} ($\pm$0.3) \\ 
\hline
\end{tabular}
\end{center}
\caption{Comparison of \ebert and \mbert on the 45-tag \posi task. We report \textit{oracle} results in this table.}
\label{table:3}
\end{table}

\begin{figure*}
	\centering
	\begin{subfigure}{0.78\columnwidth}
		\centering
		\includegraphics[width=\columnwidth]{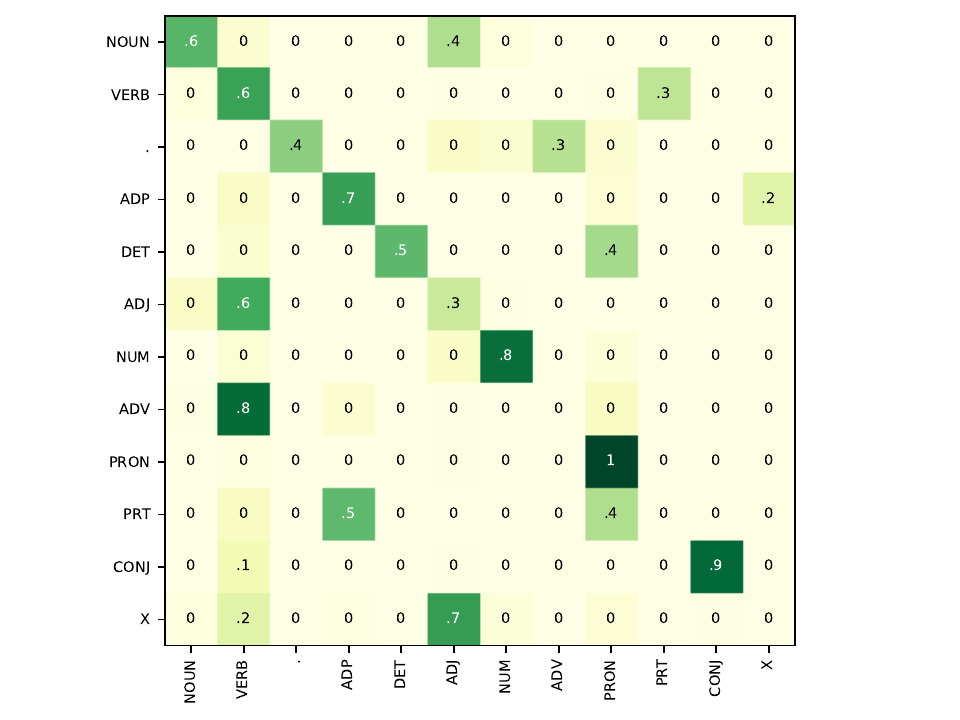}
		\caption{mBERT: One-to-One accuracy: 54.4\%}
	\end{subfigure}
	\begin{subfigure}{0.78\columnwidth}
		\centering
		\includegraphics[width=\columnwidth]{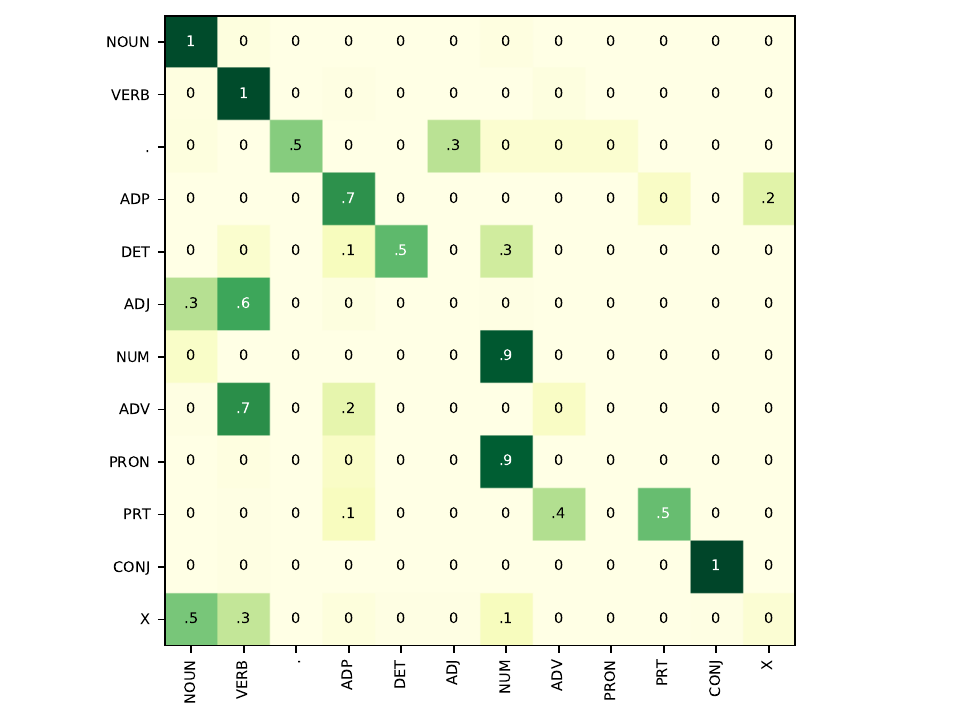}
		\caption{SyntDEC: One-to-One accuracy: 65.9\%}
	\end{subfigure}
	\caption{Comparison of confusion matrices of mBERT and SyntDEC for 12-tag experiments on English. One-to-one mapping is used to assign labels to clusters.}\label{fig:cm_12_tag_sub}
\end{figure*}

\begin{figure}
\centering
\includegraphics[width=0.7\columnwidth]{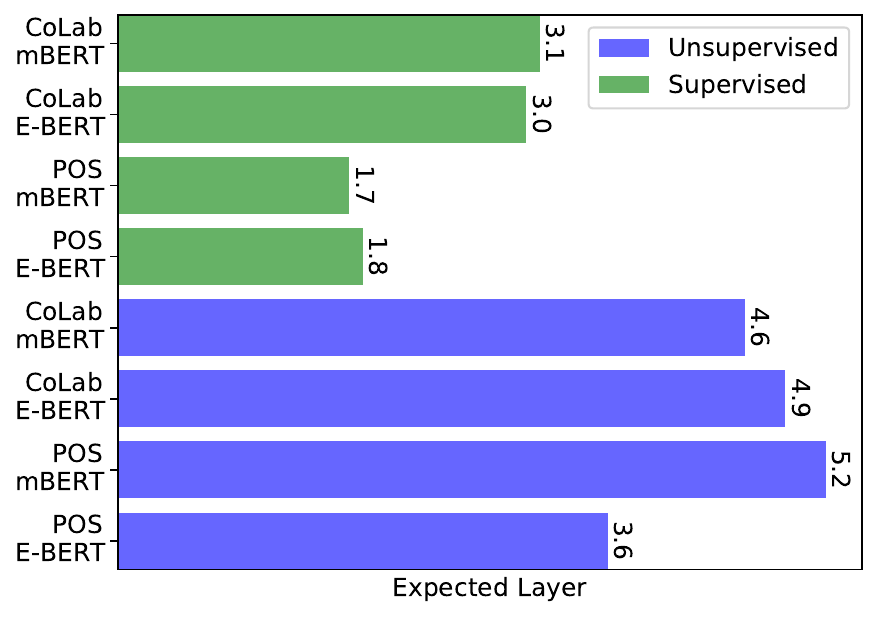}
\caption{Expected Layer of \posi and \colab under unsupervised \syntdec(blue) and supervised settings (green) with \ebert and \mbert representations.}
\label{fig:explayer}
\end{figure}

\begin{figure}
\centering
\begin{subfigure}[t]{.494\columnwidth}
\centering
\includegraphics[trim=0cm 0cm 0cm 0cm, clip=true, width=\textwidth]{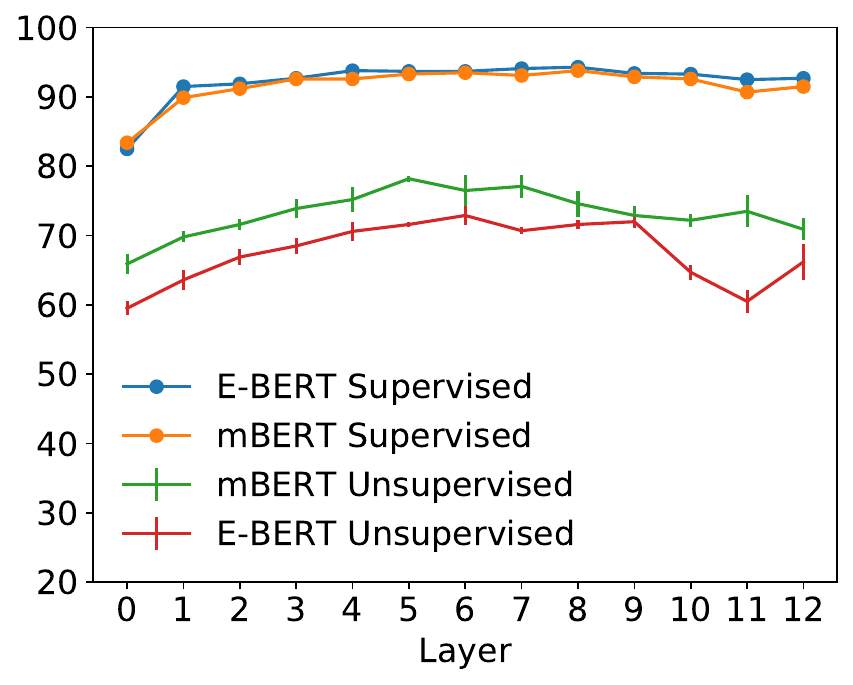}
\caption{\posi}
\label{fig:pos_layerwise}
\end{subfigure}
\begin{subfigure}[t]{.494\columnwidth}
\centering
\includegraphics[trim=0cm 0cm 0cm 0cm, clip=true, width=\textwidth]{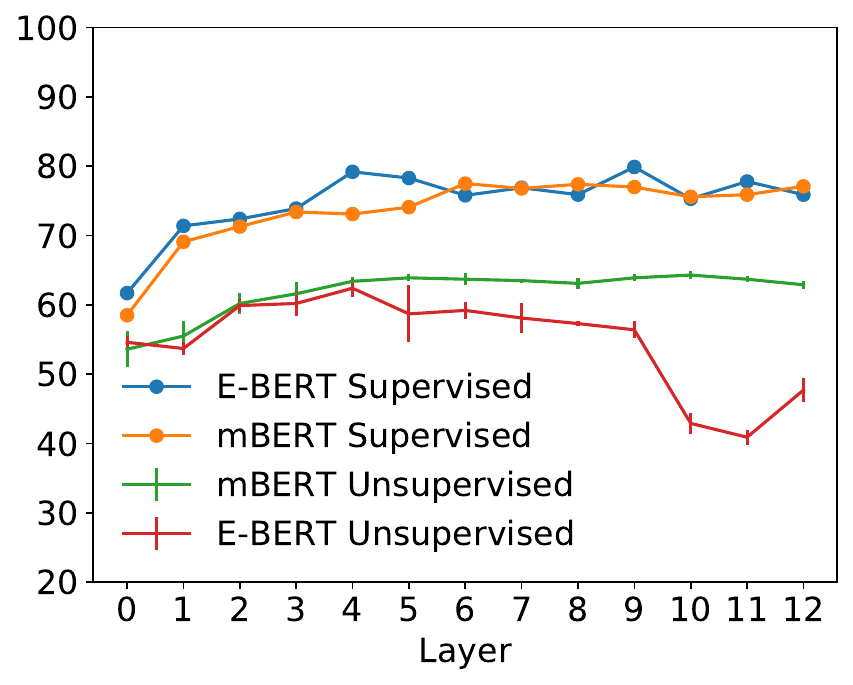}
\caption{\colab}
\label{fig:colab_layerwise}
\end{subfigure}
\caption{Comparison of M1/F1 measure for \posi and \colab under unsupervised (\syntdec) and supervised settings with \mbert and \ebert representations.}
\label{fig:layerwise}
\end{figure}

\begin{comment}
\begin{figure}
\centering
\begin{subfigure}[t]{.45\columnwidth}
\centering
\includegraphics[trim=2cm 0cm 2cm 0cm, clip=true, width=\textwidth]{sections/images/mbert_nouns.pdf}
\caption{\mbert}
\end{subfigure}
\begin{subfigure}[t]{.45\columnwidth}
\centering
\includegraphics[trim=2cm 0cm 2cm 0cm, clip=true, width=\textwidth]{sections/images/syntdec_morph_nouns.pdf}
\caption{\syntdec Morph}
\end{subfigure}
\caption{Noun CM}\label{fig:noun_cm}
\end{figure}

\begin{figure}
\centering
\begin{subfigure}[t]{.45\columnwidth}
\centering
\includegraphics[trim=2cm 0cm 2cm 0cm, clip=true, width=\textwidth]{sections/images/mbert_noun_pronoun.pdf}
\caption{\mbert}
\end{subfigure}
\begin{subfigure}[t]{.45\columnwidth}
\centering
\includegraphics[trim=2cm 0cm 2cm 0cm, clip=true, width=\textwidth]{sections/images/syntdec_noun_pronoun.pdf}
\caption{\syntdec\unskip\_morph}
\end{subfigure}
\caption{Matrix showing Noun Pronoun Confusions}\label{fig:noun_pn_cm}
\end{figure}
\end{comment}

\noindent{\bf Error Analysis:}
We compared \syntdec and \kmeansmethod (when both use \mbert) and found that \syntdec does better on noun phrases and nominal tags. 
%\vikram{Yes, it is kmeans over mbert}\kevin{What are the two methods being compared in this discussion? We seem to be comparing \syntdec and \mbert.. should the latter be ``\kmeansmethod on \mbert''?} 
It helps alleviate confusion among fine-grained noun tags (e.g., NN vs.~NNS), while also showing better handling of numerals (CD) and personal pronouns (PRP). However, \syntdec still shows considerable confusion among fine-grained verb categories. %(see confusions between VBN/VBD, VBG/VB, VBP/VB, and VBZ/VBN)%. 
For 12-tag experiments, we similarly found that \syntdec outperforms \kmeansmethod for the majority of the tags, especially nouns and verbs, resulting in a gain of more than 20\% in 1-to-1 accuracy. We further compare t-SNE visualizations of \syntdec and \mbert embeddings and observe that \syntdec embeddings show relatively compact clusters.
Detailed results and visualizations are shown in
Figure \ref{fig:cm_12_tag_sub} and the appendix.
%}
%SyntDEC performs better or maintain similar performance to mBERT for frequent noun forms. It is interesting to note that for a less frequent NNPS tag, SyntDEC has a confusion with other noun tags instead of other tags which demonstrates that embeddings of nouns are aligned closely. For pronouns, SyntDEC is able to resolve the confusion between personal and possessive pronouns.
%\vikram{Check the figure ordering later}
%We present the full confusion matrix in (supplementary;figure 4 and 5). 
%\mrinmaya{need to add more observations from Figure \ref{cm_syntdec_morph_45tag} here, if any}

%We visualize embeddings learnt by SyntDEC on the 12 tag universal treebank dataset along with gold POS tags in Figure \ref{fig:tsne_no_c}. 
%We note that SyntDEC does a relatively good job of embedding tokens into syntactically consistent clusters.
%\mrinmaya{We observe some confusions...}

\begin{table*}[h]
\scriptsize
%\footnotesize	
\begin{center}
\begin{tabular}{|l|c|c|c|c|c|c|c||c|c|c|c|} 
 \cline{3-11}
 \multicolumn{2}{c|}{} & \multicolumn{6}{c||}{\bf Nearby} & \multicolumn{3}{c|}{\bf Distant}\\
 \hline
  & {\bf en} & {\bf de} & {\bf sv} & {\bf es} & {\bf fr} & {\bf pt} & {\bf it} & {\bf ko} & {\bf id} & {\bf ja} & {\bf Mean} \\ [0.5ex] 
  %\hline
{\bf distance to en} & 0 & 0.36 & 0.4 & 0.46 & 0.46 & 0.48 & 0.50 & 0.69 & 0.71 & 0.71 & -\\ 
\hline
\hline
\multirow{2}{*}{{\bf Monolingual}} & 76.5 & 81.5 & 68.8 & 78.9 & 70.7 & 77.7 & 71.7 & 69.7 & 76.8 & 84.7 & 75.7 \\ [0.5ex]
& ($\pm$1.1) & ($\pm$1.8) & ($\pm$3.9) & ($\pm$1.9) & ($\pm$3.9) & ($\pm$2.1) & ($\pm$3.3) & ($\pm$1.5) & ($\pm$1.1) & ($\pm$1.2) & -  \\ [0.5ex]
\hline
\multirow{2}{*}{{\bf Crosslingual}} & 76.5 & 71.9 & 66.7 & 75.7 & 73.5 & 77.6 & 73.5& 67.5 & 75.4 & 80.3 & 73.9 \\ [0.5ex]
 & ($\pm$1.1) & ($\pm$1.5) & ($\pm$1.9) & ($\pm$1.4) & ($\pm$1.1) & ($\pm$1.1) & ($\pm$1.2) & ($\pm$0.9) & ($\pm$1.7) & ($\pm$1.3) & - \\ [0.5ex]
\hline
%\hline
%\multirow{2}{*}{{\bf Comblingual}} & 77.7 & 76.3 & 71.0 & 79.8 & 78.1 & 81.9 & 78.6 & 65.7 & 77.9 & 75.6 & \textbf{76.3}\\ [0.5ex]
%& ($\pm$2.4) & ($\pm$3.2) & ($\pm$3.7) & ($\pm$2.1) & ($\pm$2.8) & ($\pm$2.0) & ($\pm$2.4) & ($\pm$2.6) & ($\pm$1.9) & ($\pm$3.9) & - \\ [0.5ex]
%\hline

 \end{tabular}
 \end{center}
 \caption{\posi M1 for \syntdec with \mbert on 12-tag universal treebank in monolingual and crosslingual settings. 
 \textbf{Monolingual}: clusters are learned and evaluated on the same language. \textbf{Crosslingual}: clusters are learned on English and evaluated on all languages.
 % \textbf{Comblingual}: clusters are learned on a set of languages \{en, es, de, it\} and evaluated on all languages. 
% For crosslingual and comblingual experiments, the clusters induced on English and combination of languages are evaluated on other languages. 
%\mbert is used in all settings.
}
\label{table:6}
\end{table*}

\section{\syntdec as an Unsupervised Probe}
Next, we leverage \syntdec as an unsupervised probe to analyse where syntactic information is captured in the pretrained representations. 
%\vikram{Since we have already spoken about these things in the introduction, we can shorten it here}
Existing approaches to probing usually rely on supervised training of probes. % and use performance measures such as accuracy to understand how linguistic information is captured in these representations. 
However, as argued recently by \cite{zhou-21}, this can be unreliable.
Our supervision-free probe arguably gets rid of any bias in interpretations due to the involvement of training data in probing.%\vikram{Let us revisit the last statement. It does not look complete?}
%the contribution of layer activations for our tasks and a shallow MLP network for supervised experiments. 
We compare our unsupervised probe to a reimplementation of the supervised shallow MLP based probe in \citet{tenney-etal-2019-bert}. Similar to their paper, we report \explayer under supervised and unsupervised settings for the two tasks in Figure \ref{fig:explayer}. \explayer represents the average layer number in terms of incremental performance gains:
$
{E}_{\Delta}{[l]} = \frac{\sum_{l=1}^{L}l*\Delta^{(l)}}{\sum_{l=1}^{L}\Delta^{(l)}}
\label{eq:exp_layer}
$, where $\Delta^{(l)}$ is the change in the performance metric when adding layer $l$ to the previous layers. Layers are incrementally added from lower to higher layers. We use F1 and M1 score as the performance metric for supervised and unsupervised experiments respectively. % In Figure~\ref{fig:explayer}, we plot the \explayer under different settings. 
We observe that:
\begin{enumerate}[topsep=0pt,itemsep=-1ex,partopsep=1ex,parsep=1ex]
\item \explayer as per the unsupervised probe (blue) is higher than the supervised probe (green) for both tasks and models showing that unsupervised syntax induction benefits more from higher layers.
%K's notes:
%For unsupervised tasks, \mbert benefits more than \ebert from its higher layers.

%for sup POS, elayers are same across models, while for unsup, more diff

%Unsup probes show more difference across layers and between models than supervised. 
\item There are larger differences between \ebert and \mbert \explayer under unsupervised settings suggesting that our unsupervised probe utilizes \mbert and \ebert layers differently than the supervised one. 
\end{enumerate}

\noindent In Figure~\ref{fig:layerwise}, we further probe the performance of each layer individually by computing the F1 score for the supervised probe and the M1 score for the unsupervised probe. We observe noticeable improvement at Layer 1 for supervised \posi and Layer 1/4/6 for \colab which also correlates with their respective \explayer values. For unsupervised settings, the improvements are more evenly shared across initial layers. Although F1 and M1 are not directly comparable, 
%it is interesting to note that 
supervised performance is competitive even at higher layers while unsupervised performance drops. We present detailed results in the appendix.
\section{Crosslingual \posi}
\citet{pires2019multilingual,wu2019beto} show that \mbert is effective at zero-shot crosslingual transfer. Inspired by this, 
we evaluate the crosslingual performance on 12-tag universal treebank (Table~\ref{table:6}). 
The first row shows M1 accuracies when training and evaluating \syntdec on the same language (monolingual). The second row shows M1 accuracies of the English-trained \syntdec on other languages (crosslingual). In general, we find that clusters learned on a high-resource languages like English can be used for other languages. Similar to \citet{he2019cross}, we use the distances of the languages with English to group languages as \textit{nearby} or \textit{distant}. The distance is calculated by accounting for syntactic, genetic, and geographic distances according to the URIEL linguistic database \cite{littell2017uriel}. Our results highlight the effectiveness of \mbert in crosslingual \posi. Even for Asian languages (ko, id, and ja), which have a higher distance from English, the performance is comparable across settings. For nearby languages, crosslingual \syntdec performs well and even outperforms the monolingual setting for some languages.

% We further perform ``combilingual'' experiments where we induce clusters using a combination of European languages (en, es, it, de)\footnote{We chose  European languages for inducing clusters since we found them to transfer better to other languages (Table~\ref{cross_langs}).} and evaluate them on all languages. We observe further improvements over crosslingual experiments, suggesting that adding more data helps to obtain better clusters.

%Further study would help investigate the performance decrease for German. However, in general, these findings suggest that clusters learned on a high-resource  language like English can be used for \posi for different languages.
%\mrinmaya{What is the take-away of this experiment?}

\begin{table}[h]
\setlength{\tabcolsep}{4pt}
\centering
\resizebox{\columnwidth}{!}{
\begin{tabular}{|cl|l|c|c|c|} 
 \hline
\multicolumn{2}{|c|}{Method} & \multicolumn{1}{c|}{$F1_{\mu}$} & $F1_{\mathit{max}}\!\!$ & M1 & VM \\
\hline\hline
 %\cline{2-5}
%&Upper bound & 76.3 & 76.3 & NA\\\cline{2-5}
&\diora & 62.5 ($\pm$0.5) & 63.4  & - & -\\
&\diora$_{\!\mathit{CB}}$ (*) & 64.5 ($\pm$0.6) & 65.5  & - & -\\
%\rule{0pt}{2ex} 
&\diora$_{\!\mathit{CB}}^*$ (*) & 66.4 ($\pm$0.7) & 67.8 & - & -\\
\cline{1-6}
\parbox[t]{5.5mm}{\multirow{3}{*}{\rotatebox[origin=c]{90}{\begin{tabular}{r}\tiny{{\bf DIORA}} \\\tiny{{\bf Baselines}}\end{tabular}}}}
\rule{0pt}{2ex} 
& \ebert (**) & 41.8 & 42.2 & - & -\\
& \elmo (**) & 58.5 & 59.4 & -& -\\
\rule{0pt}{2ex} 
& \elmo$_{\!\mathit{CI}}$ (**) & 53.4 & 56.3 & - & -\\
\cline{1-6}
\parbox[t]{0.5mm}{\multirow{3}{*}{\rotatebox[origin=c]{90}{{\bf {\tiny \syntdec}}}}}
\rule{0pt}{2ex} 
&\ebert & 60.8 ($\pm$0.7) & 62.7 & 75.4 ($\pm$1.1) & 41.2 ($\pm$1.4)\\
&\spanbert & 61.3 ($\pm$0.8) & 63.3 & 75.9 ($\pm$1.0) & 40.8 ($\pm$1.1)\\
&\mbert & 64.0 ($\pm$0.4) & 64.6 & 79.6 ($\pm$0.6) & 44.5 ($\pm$0.7)\\
\hline
\end{tabular}
}
\caption{\colab results on the WSJ test set using the gold parses over five random runs. 
Our models were trained for 15 epochs and results from the final epoch for each run are recorded. 
\diora results are reported from \citet{drozdov2019unsupervised2}. $\diora_{\mathit{CB}}$ and $\diora_{\mathit{CB}}^*$ are fairly specialized models involving codebook learning (*). We also report \ebert and \elmo baselines from \citet{drozdov2019unsupervised2} (**). 
We significantly outperform these previously reported \ebert/\elmo baselines. Our results are not directly comparable to \diora as it uses the WSJ dev set for tuning and early stopping whereas we do not.}
\label{table:7}
\end{table}

\section{Constituency Labelling (\colab)}

In Table~\ref{table:7}, we report the F1 and M1 score of constituency labelling (\colab) over the WSJ test set. We represent constituents by concatenating
%\footnote{We also tried coherent \cite{seo2019real}, diffsum \cite{ouchi2018span,stern2017minimal}, and the concatenation of end words and the span mean/max but did not see substantial improvements.}
embeddings of the first and last words in the span (where word embeddings are computed by averaging corresponding subword embeddings). We observe improvement over \diora \cite{drozdov2019unsupervised1}, a recent unsupervised constituency parsing model, and achieve competitive results to recent variants that improve \diora with discrete representation learning \cite{drozdov2019unsupervised2}. %\vikram{
Our model and the \diora variants use gold constituents for these  experiments. We compute F1 metrics for comparing with previous work but also report M1 accuracies.
%} 
As with \posi, our results suggest that \mbert  outperforms both \spanbert and \ebert for the \colab task as well. We also note that \spanbert performs better than \ebert, presumably because \spanbert seeks to learn span representations explicitly. In the Appendix(Table~\ref{table:8}), we explore other ways of representing constituents and note that mean/max pooling followed by clustering does not perform well. Compressing and finetuning the mean-pooled representation using \syntdec (\syntdecmean) is also suboptimal. We hypothesize that mean/max pooling results in a loss of information about word order in the constituent whereas the concatenation of first and last words retains this information. Even a stacked autoencoder (\sae) over the concatenation of first and last token achieves competitive results, but finetuning with \syntdec improves the $F1_{\mu}$ by nearly 4.5\%. This demonstrates that for \colab also, the transformation to lower dimensions and finetuning to clustering friendly spaces is important for achieving competitive performance.

\section{Related Work}
\label{sec:related_work}
\noindent{\bf Deep Clustering:}
Unlike previous work where feature extraction and clustering were applied sequentially, deep clustering aims to jointly optimize for both by combining a clustering loss with the feature extraction. A number of deep clustering methods have been proposed which primarily differ in their clustering approach:
%\footnote{\citet{aljalbout2018clustering} and \citet{min2018survey} provide comprehensive summaries of deep clustering approaches.} 
\citet{yang2017towards} use KMeans, \citet{xie2016unsupervised} use cluster assignment hardening, \citet{ghasedi2017deep} add a balanced assignments loss on top of cluster assignment hardening, \citet{huang2014deep} introduce a locality-preserving loss and a group sparsity loss on the clustering, \citet{yang2016joint} use agglomerative clustering, and \citet{ji2017deep} use subspace clustering.
%\cite{jiang2016variational}
%\cite{caron2018deep}
%Deep clustering approaches has shown great promise in unsupervised clustering. However, m
%In principle, 
All of these approaches can be used to cluster contextualized representations, and future work may improve upon our results by exploring these approaches. 
The interplay between deep clustering for syntax and recent advancements in NLP, such as contextualized representations, has not previously been studied. In this paper, we fill this gap.

\noindent{\bf Unsupervised Syntax Induction:}
There has been a lot of work on unsupervised induction of syntax, namely, unsupervised constituency parsing \citep{klein-manning-2002-generative,seginer-2007-fast,kim2019compound} and dependency parsing \citep{klein2004corpus,smith-eisner-2006-annealing,gillenwater-etal-2010-sparsity,spitkovsky-etal-2013-breaking,jiang-etal-2016-unsupervised}. 
%In addition, recent work has studied inducing constituency parse trees based on downstream tasks \citep{choi2018learning,li2019imitation} or cross-modal supervision \citep{shi2019visually,zhao2020visually}.
While most prior work focuses on inducing \emph{unlabeled} syntactic structures, we focus on inducing constituent labels while assuming the gold syntactic structure is available. This goal has also been pursued in prior work \citep{drozdov2019unsupervised2,jin2020importance}. 
%There has also been work on inducing labels for nonterminal nodes in constituency parse trees \citep{drozdov2019unsupervised2,jin2020importance}, of which the goal aligns with the CoLab task we investigated. 
Compared to them, we present simpler models to induce syntactic labels directly from pretrained models via dimensionality reduction and clustering. Similar to us, ~\cite{li2019specializing} also note gains for supervised NLP tasks upon reducing the representation dimension.

%, which can also be easily applied for other syntactic label induction tasks. 

\noindent{\bf Probing Pretrained Representations:} Recent analysis work \citep[][\emph{inter alia}]{liu-etal-2019-linguistic,tenney2019you,aljalbout2018clustering,jawahar-etal-2019-bert} has shown that pretrained language models
%has shown that such models, e.g., ELMo \citep{peters2018deep} and BERT \citep{devlin2018bert}, are able to 
encode syntactic information efficiently. 
Most of them train a supervised model using pretrained representations and labeled examples, and show that pretrained language models effectively encode part-of-speech and constituency information. 
In contrast to these works, we propose an unsupervised approach to probing which does not rely on any training data. \cite{zhou-21} also pursue the same goals by studying the geometry of these representations.

%\kevin{I think it would be good to cite more of the related Bertology work, especially work that explicitly looks at POS (e.g., BERT rediscovers classical NLP pipeline, Nelson Liu's paper, etc.) and constituent labels.. I think our span paper has some of that related work, but not all. } 

\section{Conclusion}
In this work, we explored the problem of clustering text representations for model interpretation and induction of syntax. We observed that
%while these representations are extremely powerful, 
%clustering these representation using 
off-the-shelf methods like KMeans are sub-optimal as these representations are high dimensional and, thus, not directly suitable for clustering. Thus, we proposed a deep clustering approach which jointly transforms these representations into a lower-dimensional 
cluster friendly space 
and clusters them. Upon integration of a small number of task-specific features,
%like morphological representations and token context
and use of multilingual representations, we find that our approach achieves competitive performance %across various languages and tasks
%, establishing strong baselines 
for unsupervised \posi and \colab% with contextualized representations
comparable to more complex methods in the literature. Finally, we also show that we can use the technique as a supervision-free approach to probe syntax in these representations and contrast our unsupervised probe with supervised ones.

{
% Use \bibliography{yourbibfile} instead or the References section will not appear in your paper
%\nobibliography{aaai22}
\bibliography{iqa}
}
%\section{Acknowledgments}

\clearpage

\appendix
\section{Task and Architecture}
\label{sec:appendix}
\begin{comment}
\subsection{Task Illustration}
\label{appendix:task_illustration}
\begin{figure}
\centering
\includegraphics[width=\columnwidth]{sections/images/Braces-Example-POS-COLAB.pdf}
\caption{
An illustration of the \posi and \colab tasks.
}
\label{pos_colab}
\end{figure}
Figure~\ref{pos_colab} shows an example of \posi and \colab. Both tasks depend on the context and motivate us to leverage contextualized embeddings.
\end{comment}
\subsection{\cbow Illustration}
\label{cbow_illustration}
\begin{figure}
\centering
\includegraphics[height=6 cm, keepaspectratio]{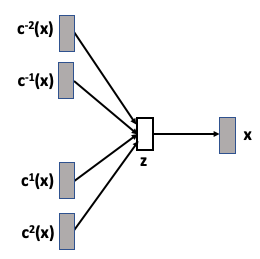}
\caption{\cbow variant of \syntdec.  Embeddings of the context tokens are concatenated and used as input to \syntdec to reconstruct the embedding of the token. 
}\label{fig:syntdec-cbow}

\end{figure}

Figure~\ref{fig:syntdec-cbow} shows an illustration of \syntdec--\cbow
\begin{comment}

\begin{figure}
\centering
\begin{subfigure}[t]{.485\columnwidth}
\centering
\includegraphics[width=\textwidth]{sections/images/mbert_newcolor.jpg}
\caption{\mbert}
\end{subfigure}
\begin{subfigure}[t]{.485\columnwidth}
\centering
\includegraphics[width=\textwidth]{sections/images/syntdec_newcolor.pdf}
\caption{\syntdec}
\end{subfigure}
\caption{t-SNE visualization of \mbert and \syntdec embeddings of tokens from the Penn Treebank. Colors correspond to the ground truth POS tags (best viewed in color).}\label{fig:tsne}
\end{figure}
\end{comment}

\begin{figure*}
\centering
\includegraphics[trim=1cm 0cm 2cm 0cm, clip=true, width=0.9\columnwidth]{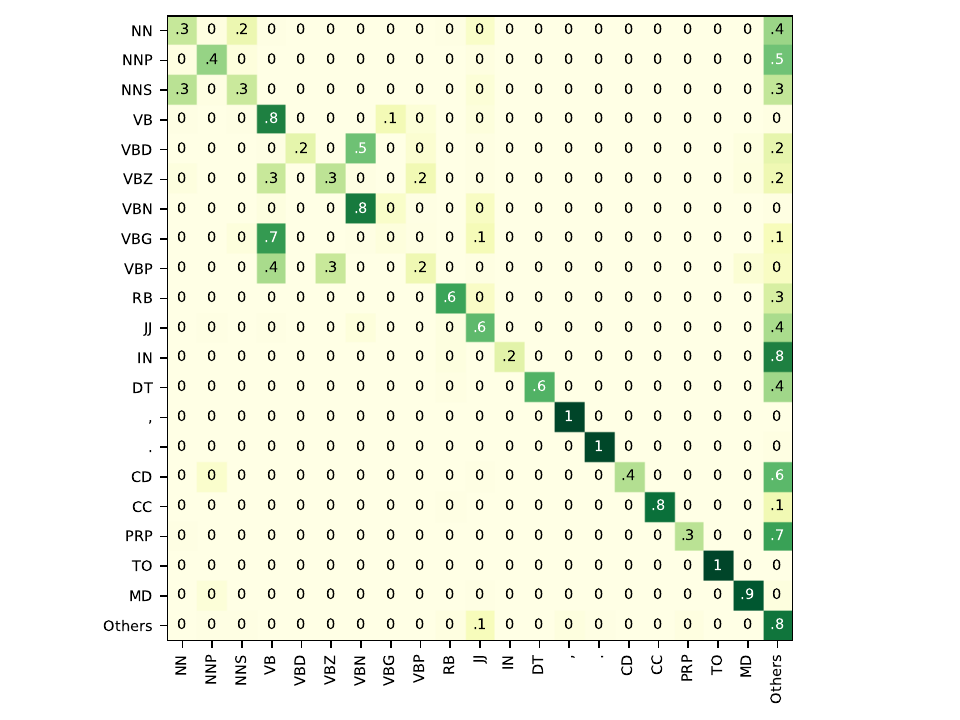}
\includegraphics[trim=1cm 0cm 2cm 0cm, clip=true, width=0.9\columnwidth]{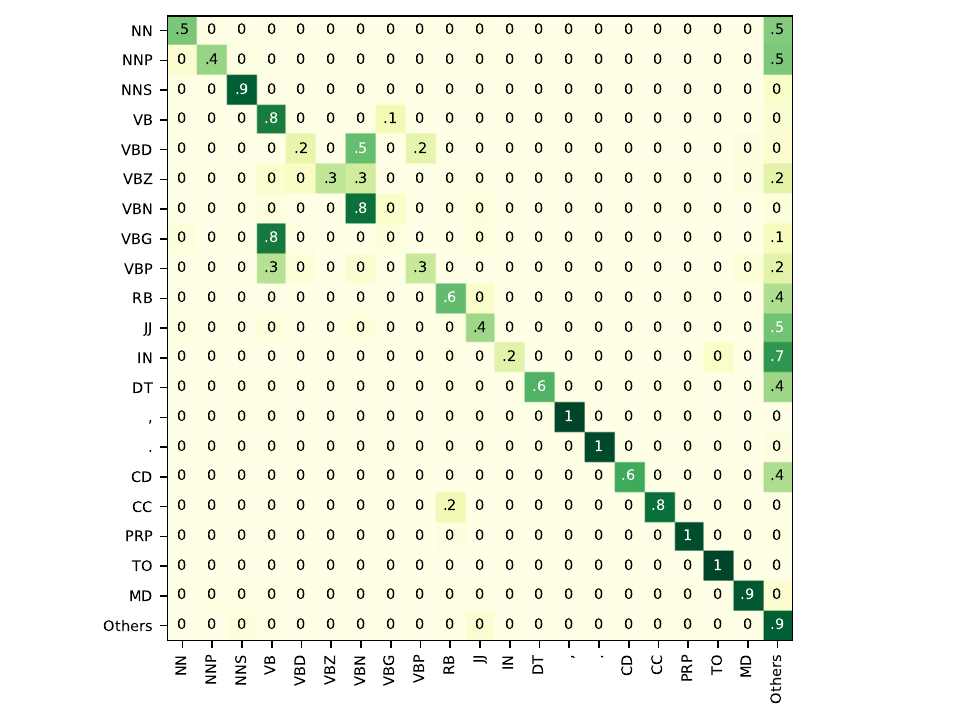}
\caption{
Confusion matrices for \kmeansmethod over \mbert (left) and \syntdecmorph (right) for the 20 most frequent tags (45-tag \posi).
}
\label{cm_syntdec_morph_45tag}
\end{figure*}

\section{\posi Analysis}
\label{section:posi_analysis}
\subsection{45-Tag \posi Analysis}
\label{section:45_Tag_Universal Treebank_Analysis}
We show a t-SNE visualization of \mbert embeddings and the embeddings learned by our deep clustering model in Figure \ref{fig:tsne}. We note that the clusters formed by \syntdec are more coherent and dense.

In Figure \ref{cm_syntdec_morph_45tag}, we show the confusion matrices of \syntdecmorph and \mbert for the 20 most frequent tags in the 45 tag \posi task by assigning labels to predicted clusters using the optimal 1-to-1 mapping. We observe that \syntdecmorph outperforms \mbert for most tags. 

\subsection{12-Tag \posi Analysis}
\label{section:12_Tag_Universal Treebank_Analysis}
\begin{figure*}
\centering
\begin{subfigure}[t]{\columnwidth}
\centering
\includegraphics[width=\textwidth]{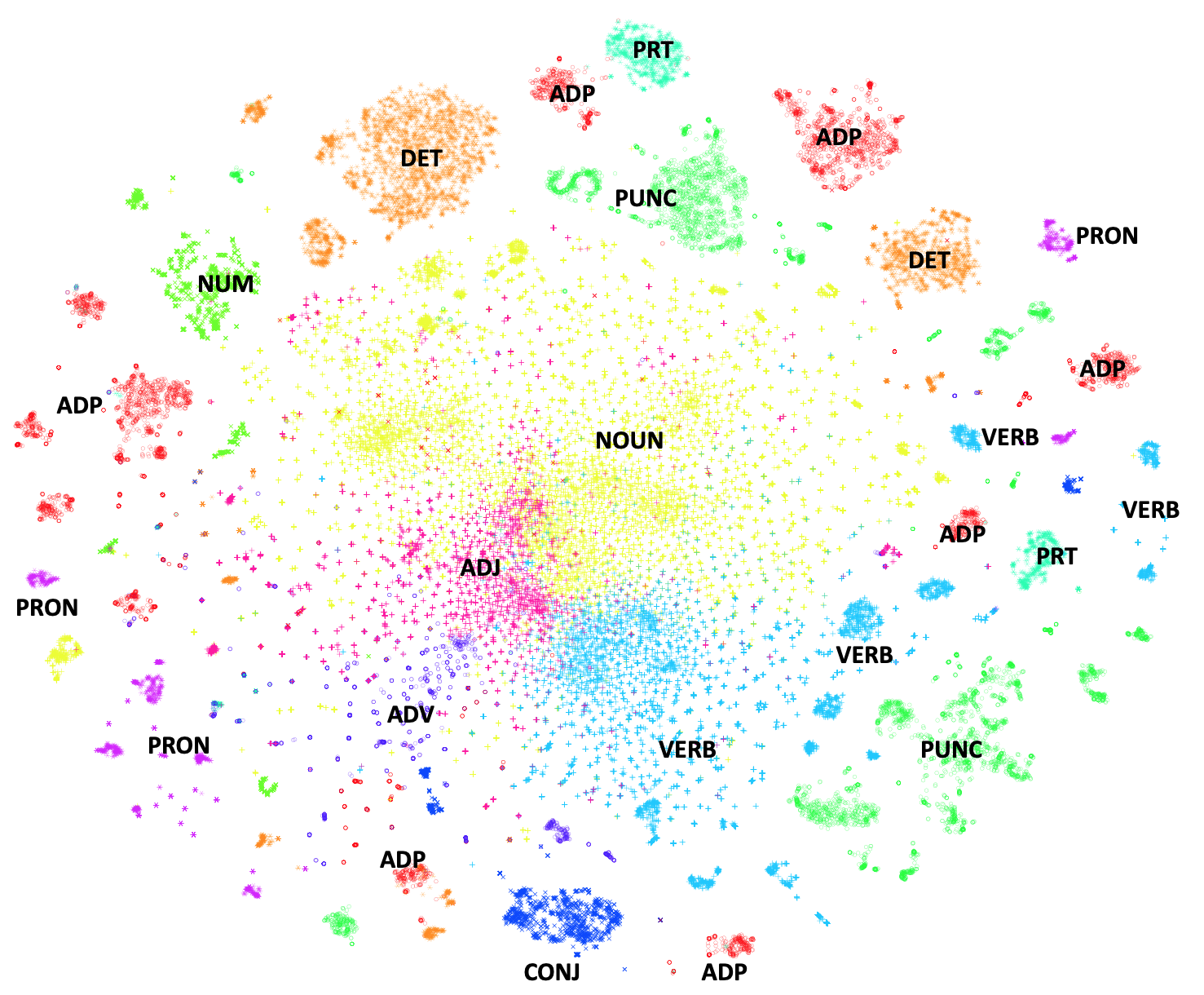}
\caption{\mbert embeddings}
\end{subfigure}
\begin{subfigure}[t]{\columnwidth}
\centering
\includegraphics[width=\textwidth]{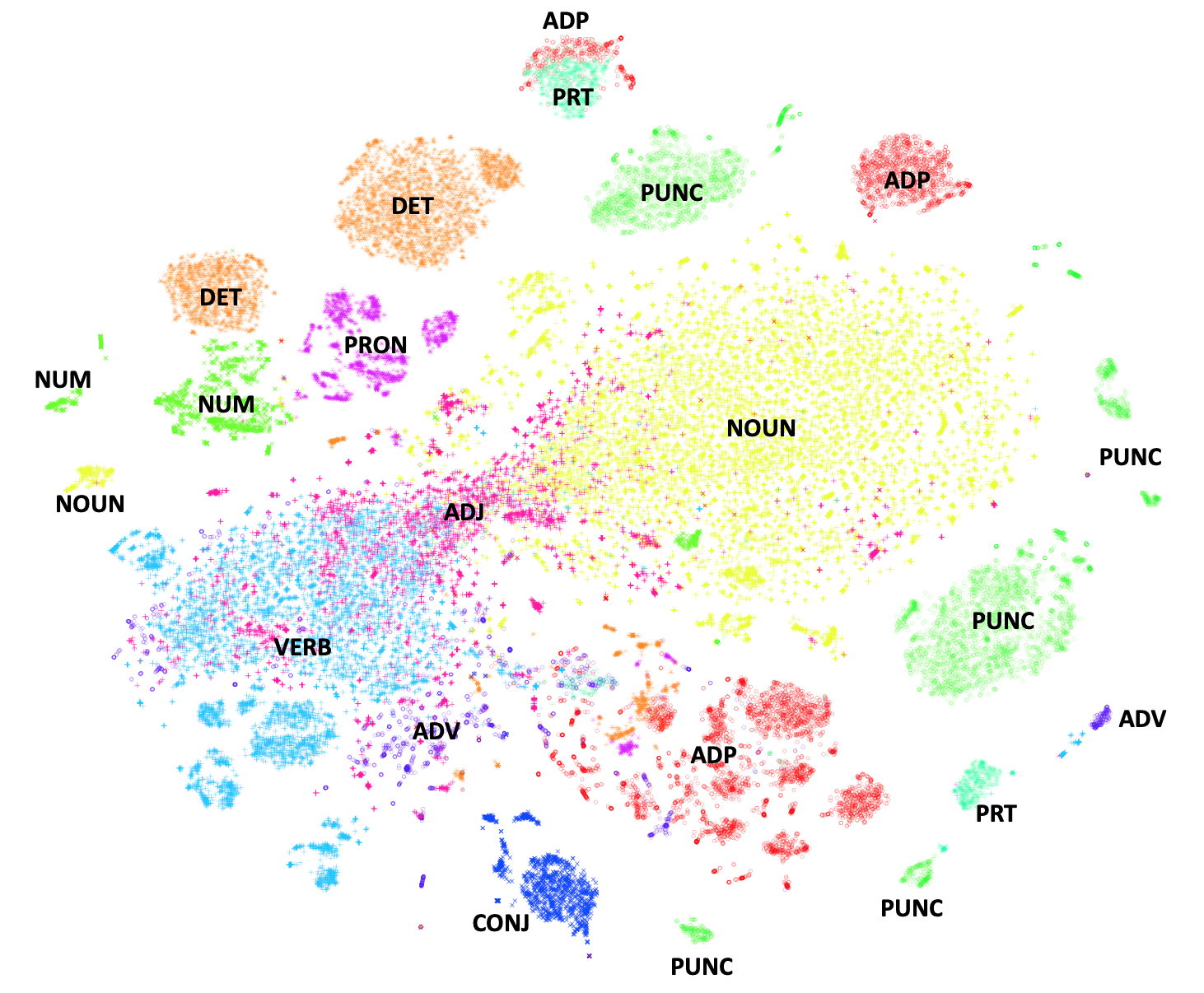}
\caption{\syntdec embeddings}
\end{subfigure}
\caption{t-SNE visualization of \mbert and \syntdec embeddings of tokens from 12 tag Universal Treebank english dataset. Colors correspond to the ground truth POS tags.}\label{fig:tsne_no_c}
\end{figure*}

In Fig~\ref{fig:tsne_no_c}, we show t-SNE visualization of \syntdec and \mbert embeddings of tokens from the 12-tag Universal Treebank English dataset. \syntdec embeddings produce more distinct clusters. 

\subsection{45-Tag \posi Ablation Studies}
\label{section:45_Tag_Ablation_Studies}
\begin{table}[h]
\begin{center}
%\small
\begin{tabular}{ |l|c|c| } 
 \hline
 Morph. Feats. & M1 & VM\\ [0.5ex] 
 \hline\hline
Unigram & 77.9 ($\pm$2.3) & 73.2 ($\pm$1.4)\\
Bigram & 77.4 ($\pm$2.4) & 72.7 ($\pm$1.9)\\
Trigram & {\bf 79.5} ($\pm$0.9)  & 73.9 ($\pm$0.7)\\
\hline
\end{tabular}
\end{center}
\caption{Comparison of different orders of character $n$-gram  embeddings for the 45-tag \posi task.}
\label{table:4}
\end{table}
In Table~\ref{table:4}, we study the impact of different character embeddings and achieve best results on using embeddings of the trailing trigram of each token. 

\subsection{\colab Ablation Studies}
\label{section:Colab_Ablation_Studies}
In Table~\ref{table:8}, we present the ablation results for \colab. We find that \kmeansmethod on Max or mean pooled span representation of \mbert do not work well. Even deep clustering (\syntdecmean) over the mean of the span representation does not help. \sae and \syntdec trained over the concatenation of the representation of the end points  substantially improve the results.
\begin{table}
\small
\setlength{\tabcolsep}{5.4pt}
\begin{center}
\begin{tabular}{|l|c|c|c|} 
 \hline
  Method & $F1_{\mu}$ & $F1_{\mathit{max}}$ & M1\\
 \hline\hline
 \kmeansmethod(Mean) & 39.9 ($\pm$0.4) & 40.2 & 48.6 ($\pm$0.3) \\
 \kmeansmethod(Max) & 40.1 ($\pm$0.6) & 40.9 & 49.6 ($\pm$0.7) \\
 \syntdecmean & 40.8 ($\pm$1.1) & 42.4 & 49.8 ($\pm$1.2) \\
 \hline
 \hline
 \sae\ & 61.2 ($\pm$1.2) & 62.8 & 76.6 ($\pm$1.1) \\
 \syntdec & 64.0 ($\pm$0.4) & 64.6 & 79.6 ($\pm$0.6) \\
 \hline
\end{tabular}
\caption{Comparison of different methods to represent spans for \colab. \mbert is used in these experiments.}
\label{table:8}
\end{center}
\end{table}

\section{Unsupervised Probing}
\label{section:unsupervised_probing}
In Table~\ref{table:ebert_pos_inc_layer_wise} and Table~~\ref{table:mbert_pos_inc_layer_wise}, we report the results of adding layers incrementally from lower to higher for \posi on \mbert and \ebert. We present similar results for \colab in Table~\ref{table:ebert_colab_inc_layer_wise} and Table~\ref{table:mbert_colab_inc_layer_wise}. In Table~\ref{table:ebert_posi_layer_wise} and Table~~\ref{table:mbert_posi_layer_wise}, we report the results of individual layers for \posi on \mbert and \ebert. We present similar results for \colab in Table~\ref{table:ebert_colab_layer_wise} and Table~\ref{table:mbert_colab_layer_wise}.

\section{Hyperparameters}
\label{hyperparameters}
Words are represented by 768 dimension vectors obtained after taking the mean of BERT layers.We tried max and mean pooling also but did not notice much improvement. Morphological embeddings extracted from \fasttext have 300 dimensions. 
The number of clusters are set equal to the number of ground truth tags for all the experiments.
Following previous work \cite{stratos2018mutual}, we use the 45-tag \posi experiments on English to select the hyperparameters for our framework and use these hyperparameters across all the other languages and tasks. 

We use a \syntdec architecture with one encoder layer and use 75 as the size of the latent dimension.
%\footnote{We did not observe much improvement by increasing the number of layers or by using non-linearity and batch normalization.}
Layer-wise and end-to-end training is done for 50 epochs with a batch size of 64, learning rate of 0.1 and momentum of 0.9 using the SGD optimizer. In the clustering stage, we train \syntdec for 4000 iterations with 256 as batch size
and 0.001 as learning rate with 0.9 momentum using SGD. We set the reconstruction error weight $\lambda =5$ for all our experiments. We set the context width as one for \cbow.
For out-of-vocabulary words, we use an average over all subword embeddings. For all the experiments, we report results for the last training iteration as we do not have access to the ground truth labels for model selection. For the supervised experiments, we follow the training and architecture details of~\cite{tenney-etal-2019-bert}. All our experiments are performed on a 12GB GeForce RTX 2080 Ti GPU and each run takes approximately 3 hours.

\begin{table}[h]
\Large
\begin{center}
\begin{tabular}{ |c|c|c| } 
 \hline
 Layers & M1 & VM\\ [0.5ex] 
 \hline\hline
Layer 0 & {61.6} ($\pm$0.5) & 59.8 ($\pm$0.6) \\
Layer 0\_1 & {61.9} ($\pm$0.8) & 59.9 ($\pm$0.8) \\
Layer 0\_2 & {66.5} ($\pm$1.0) & 64.5 ($\pm$1.0) \\
Layer 0\_3 & {67.4} ($\pm$2.4) & 65.6 ($\pm$1.6) \\
Layer 0\_4 & {68.5} ($\pm$2.2) & 65.9 ($\pm$1.6) \\
Layer 0\_5 & {69.4} ($\pm$2.4) & 66.2 ($\pm$1.3) \\
Layer 0\_6 & {70.7} ($\pm$1.2) & 67.1 ($\pm$1.6) \\
Layer 0\_7 & {72.8} ($\pm$1.2) & 68.3 ($\pm$0.7) \\
Layer 0\_8 & {72.6} ($\pm$0.6) & 68.6 ($\pm$0.3) \\
Layer 0\_9 & {72.7} ($\pm$0.7) & 68.9 ($\pm$0.5) \\
Layer 0\_10 & {72.1} ($\pm$1.4) & 67.9 ($\pm$0.9) \\
Layer 0\_11 & {72.0} ($\pm$1.2) & 67.9 ($\pm$0.9) \\
Layer 0\_12 & {72.7} ($\pm$1.2) & 68.9 ($\pm$0.8) \\
\hline
\end{tabular}
\end{center}
\caption{Comparison of different \ebert layers for the 45-tag \posi. We report \textit{oracle} M1 accuracy and V-Measure (VM) averaged over 5 random runs.}
\label{table:ebert_pos_inc_layer_wise}
\end{table}

\begin{table}[h]
\Large
\begin{center}
\begin{tabular}{ |c|c|c| } 
 \hline
 Layers & M1 & VM\\ [0.5ex] 
 \hline\hline
Layer 0 & {69.6} ($\pm$2.7) & 66.4 ($\pm$2.0) \\
Layer 0\_1 & {69.8} ($\pm$1.9) & 66.9 ($\pm$0.7) \\
Layer 0\_2 & {72.1} ($\pm$1.7) & 68.2 ($\pm$0.9) \\
Layer 0\_3 & {71.5} ($\pm$1.6) & 68.5 ($\pm$0.9) \\
Layer 0\_4 & {72.1} ($\pm$1.7) & 68.5 ($\pm$0.8) \\
Layer 0\_5 & {73.1} ($\pm$1.5) & 69.1 ($\pm$0.8) \\
Layer 0\_6 & {75.0} ($\pm$1.7) & 70.1 ($\pm$1.6) \\
Layer 0\_7 & {76.2} ($\pm$2.6) & 71.5 ($\pm$1.8) \\
Layer 0\_8 & {77.9} ($\pm$1.3) & 72.2 ($\pm$1.1) \\
Layer 0\_9 & {77.8} ($\pm$1.9) & 72.6 ($\pm$1.0) \\
Layer 0\_10 & {76.9} ($\pm$2.8) & 72.1 ($\pm$1.8) \\
Layer 0\_11 & {77.5} ($\pm$0.9) & 72.1 ($\pm$0.6) \\
Layer 0\_12 & {77.8} ($\pm$1.4) & 72.6 ($\pm$1.0) \\
\hline
\end{tabular}
\end{center}
\caption{Comparison of different \mbert layers for the 45-tag \posi task. We report \textit{oracle} M1 accuracy and V-Measure (VM) averaged over 5 random runs.}
\label{table:mbert_pos_inc_layer_wise}
\end{table}

\begin{table}[h]
\Large
\begin{center}
\begin{tabular}{ |c|c|c| } 
 \hline
 Layers & M1 & VM\\ [0.5ex] 
 \hline\hline
Layer 0 & {54.3} ($\pm$1.5) & 31.6 ($\pm$1.7) \\
Layer 0\_1 & {54.2} ($\pm$1.9) & 31.5 ($\pm$2.0) \\
Layer 0\_2 & {52.6} ($\pm$1.3) & 32.6 ($\pm$2.2) \\
Layer 0\_3 & {58.8} ($\pm$0.7) & 37.2 ($\pm$1.1) \\
Layer 0\_4 & {58.9} ($\pm$1.8) & 38.0 ($\pm$2.4) \\
Layer 0\_5 & {61.3} ($\pm$0.1) & 42.1 ($\pm$0.1) \\
Layer 0\_6 & {60.5} ($\pm$1.9) & 40.7 ($\pm$3.0) \\
Layer 0\_7 & {62.1} ($\pm$0.7) & 42.7 ($\pm$1.0) \\
Layer 0\_8 & {60.1} ($\pm$2.9) & 40.7 ($\pm$3.5) \\
Layer 0\_9 & {61.9} ($\pm$0.2) & 42.7 ($\pm$1.0) \\
Layer 0\_10 & {60.8} ($\pm$1.7) & 42.0 ($\pm$2.3) \\
Layer 0\_11 & {60.6} ($\pm$3.6) & 41.9 ($\pm$4.4) \\
Layer 0\_12 & {61.4} ($\pm$0.5) & 42.4 ($\pm$0.8) \\
\hline
\end{tabular}
\end{center}
\caption{Comparison of different \ebert layers for \colab task. We report \textit{oracle} M1 accuracy and V-Measure (VM) averaged over 5 random runs.}
\label{table:ebert_colab_inc_layer_wise}
\end{table}

\begin{table}[h]
\Large
\begin{center}
\begin{tabular}{ |c|c|c| } 
 \hline
 Layers & M1 & VM\\ [0.5ex] 
 \hline\hline
Layer 0 & {54.0} ($\pm$2.1) & 32.9 ($\pm$1.7) \\
Layer 0\_1 & {53.9} ($\pm$3.3) & 33.6 ($\pm$2.5) \\
Layer 0\_2 & {58.4} ($\pm$1.7) & 36.3 ($\pm$1.4) \\
Layer 0\_3 & {56.8} ($\pm$3.2) & 35.9 ($\pm$2.3) \\
Layer 0\_4 & {60.0} ($\pm$1.8) & 39.1 ($\pm$1.7) \\
Layer 0\_5 & {61.2} ($\pm$1.3) & 40.4 ($\pm$1.5) \\
Layer 0\_6 & {62.9} ($\pm$0.6) & 43.1 ($\pm$0.5) \\
Layer 0\_7 & {62.7} ($\pm$0.7) & 42.5 ($\pm$1.1) \\
Layer 0\_8 & {62.9} ($\pm$0.4) & 43.1 ($\pm$0.6) \\
Layer 0\_9 & {62.8} ($\pm$0.6) & 42.9 ($\pm$0.9) \\
Layer 0\_10 & {63.2} ($\pm$0.5) & 43.5 ($\pm$0.6) \\
Layer 0\_11 & {63.3} ($\pm$0.6) & 43.5 ($\pm$0.8) \\
Layer 0\_12 & {64.2} ($\pm$0.4) & 45.0 ($\pm$0.7) \\
\hline
\end{tabular}
\end{center}
\caption{Comparison of different \mbert layers for \colab task. We report \textit{oracle} M1 accuracy and V-Measure (VM) averaged over 5 random runs.}
\label{table:mbert_colab_inc_layer_wise}
\end{table}

\begin{table}[h]
\Large
\begin{center}
\begin{tabular}{ |c|c|c| } 
 \hline
 Layers & M1 & VM\\ [0.5ex] 
 \hline\hline
Layer 0 & {54.6} ($\pm$0.9) & 31.8 ($\pm$1.0) \\
Layer 1 & {53.7} ($\pm$0.9) & 33.6 ($\pm$2.0) \\
Layer 2 & {59.9} ($\pm$1.0) & 39.7 ($\pm$1.0) \\
Layer 3 & {60.2} ($\pm$1.8) & 40.8 ($\pm$2.2) \\
Layer 4 & {62.4} ($\pm$1.3) & 44.0 ($\pm$1.8) \\
Layer 5 & {58.7} ($\pm$4.1) & 39.3 ($\pm$5.5) \\
Layer 6 & {59.2} ($\pm$1.2) & 39.1 ($\pm$1.9) \\
Layer 7 & {58.1} ($\pm$2.1) & 36.9 ($\pm$2.6) \\
Layer 8 & {57.3} ($\pm$0.4) & 37.5 ($\pm$0.8) \\
Layer 9 & {56.4} ($\pm$1.2) & 34.9 ($\pm$1.4) \\
Layer 10 & {42.9} ($\pm$1.5) & 16.43 ($\pm$2.3) \\
Layer 11 & {40.9} ($\pm$1.1) & 14.9 ($\pm$1.2) \\
Layer 12 & {47.7} ($\pm$1.7) & 22.3 ($\pm$2.0) \\
\hline
\end{tabular}
\end{center}
\caption{Comparison of different \ebert layers for \colab task. We report \textit{oracle} M1 accuracy and V-Measure (VM) averaged over 5 random runs.}
\label{table:ebert_colab_layer_wise}
\end{table}
\begin{table}[h]
\Large
\begin{center}
\begin{tabular}{ |c|c|c| } 
 \hline
 Layers & M1 & VM\\ [0.5ex] 
 \hline\hline
Layer 0 & {53.6} ($\pm$2.6) & 32.1 ($\pm$1.9) \\
Layer 1 & {55.5} ($\pm$2.2) & 35.3 ($\pm$1.4) \\
Layer 2 & {60.2} ($\pm$1.5) & 39.1 ($\pm$0.9) \\
Layer 3 & {61.6} ($\pm$1.7) & 41.2 ($\pm$0.9) \\
Layer 4 & {63.4} ($\pm$0.6) & 43.0 ($\pm$0.7) \\
Layer 5 & {63.9} ($\pm$0.5) & 44.1 ($\pm$0.6) \\
Layer 6 & {63.7} ($\pm$0.9) & 44.7 ($\pm$0.9) \\
Layer 7 & {63.5} ($\pm$0.3) & 44.9 ($\pm$0.5) \\
Layer 8 & {63.1} ($\pm$0.8) & 44.9 ($\pm$0.8) \\
Layer 9 & {63.9} ($\pm$0.5) & 46.2 ($\pm$0.8) \\
Layer 10 & {64.3} ($\pm$0.6) & 46.3 ($\pm$0.8) \\
Layer 11 & {63.7} ($\pm$0.4) & 45.1 ($\pm$0.2) \\
Layer 12 & {62.9} ($\pm$0.6) & 43.3 ($\pm$0.6) \\
\hline
\end{tabular}
\end{center}
\caption{Comparison of different \mbert layers for \colab task. We report \textit{oracle} M1 accuracy and V-Measure (VM) averaged over 5 random runs.}
\label{table:mbert_colab_layer_wise}
\end{table}

\begin{table}[h]
\Large
\begin{center}
\begin{tabular}{ |c|c|c| } 
\hline
Layers & M1 & VM\\ [0.5ex] 
\hline\hline
Layer 0 & {66.9} ($\pm$1.4) & 64.5 ($\pm$0.6) \\
Layer 1 & {70.8} ($\pm$0.8) & 67.4 ($\pm$0.4) \\
Layer 2 & {72.6} ($\pm$0.8) & 68.1 ($\pm$0.4) \\
Layer 3 & {74.9} ($\pm$1.4) & 70.0 ($\pm$1.0) \\
Layer 4 & {76.2} ($\pm$1.8) & 71.3 ($\pm$1.3) \\
Layer 5 & {79.2} ($\pm$0.4) & 72.9 ($\pm$0.6) \\
Layer 6 & {77.5} ($\pm$2.3) & 72.0 ($\pm$1.4) \\
Layer 7 & {78.1} ($\pm$1.7) & 71.7 ($\pm$1.1) \\
Layer 8 & {75.6} ($\pm$1.9) & 70.0 ($\pm$1.5) \\
Layer 9 & {73.9} ($\pm$1.3) & 68.5 ($\pm$0.3) \\
Layer 10 & {73.2} ($\pm$0.9) & 69.1 ($\pm$0.5) \\
Layer 11 & {74.5} ($\pm$2.3) & 69.6 ($\pm$1.1) \\
Layer 12 & {71.9} ($\pm$1.6) & 66.3 ($\pm$1.1) \\
\hline
\end{tabular}
\end{center}
\caption{Comparison of different \mbert layers for the 45-tag \posi task. We report \textit{oracle} M1 accuracy and V-Measure (VM) averaged over 5 random runs.}
\label{table:mbert_posi_layer_wise}
\end{table}

\begin{table}[h]
\Large
\begin{center}
\begin{tabular}{ |c|c|c| } 
 \hline
 Layers & M1 & VM\\  [0.5ex] 
 \hline\hline
Layer 0 & {60.5} ($\pm$1.0) & 60.0 ($\pm$0.7) \\
Layer 1 & {64.6} ($\pm$1.5) & 64.3 ($\pm$0.8) \\
Layer 2 & {67.9} ($\pm$1.2) & 66.2 ($\pm$0.4) \\
Layer 3 & {69.5} ($\pm$1.1) & 66.6 ($\pm$0.7) \\
Layer 4 & {71.6} ($\pm$1.4) & 67.2 ($\pm$0.9) \\
Layer 5 & {72.6} ($\pm$0.4) & 68.1 ($\pm$0.6) \\
Layer 6 & {73.9} ($\pm$1.4) & 67.9 ($\pm$0.9) \\
Layer 7 & {71.7} ($\pm$0.5) & 67.2 ($\pm$0.5) \\
Layer 8 & {72.6} ($\pm$0.6) & 67.2 ($\pm$0.5) \\
Layer 9 & {73.0} ($\pm$0.9) & 67.7 ($\pm$0.7) \\
Layer 10 & {65.7} ($\pm$1.1) & 59.9 ($\pm$0.6) \\
Layer 11 & {61.5} ($\pm$1.7) & 54.4 ($\pm$1.0) \\
Layer 12 & {67.2} ($\pm$2.6) & 60.4 ($\pm$2.2) \\
\hline
\end{tabular}
\end{center}
\caption{Comparison of different \ebert layers for the 45-tag \posi task. We report \textit{oracle} M1 accuracy averaged and V-Measure (VM) over 5 random runs.}
\label{table:ebert_posi_layer_wise}
\end{table}

\end{document}